\documentclass{article}

\usepackage{microtype}
\usepackage{graphicx}
\usepackage{subfigure}
\usepackage{booktabs} 
\usepackage{xcolor}
\usepackage{colortbl}
\usepackage{amsmath}
\usepackage{amssymb}
\usepackage{bm}
\usepackage{amsfonts} 
\usepackage{makecell}
\usepackage{multirow}
\usepackage{multicol}
\usepackage{amsthm}
\usepackage{soul}
\usepackage{pifont}
\usepackage{hyperref}

\usepackage[accepted]{icml2023}

\usepackage{amsmath}
\usepackage{amssymb}
\usepackage{mathtools}
\usepackage{amsthm}
\usepackage{soul}

\usepackage[capitalize,noabbrev]{cleveref}

\theoremstyle{plain}

\theoremstyle{definition}

\theoremstyle{remark}

\usepackage[textsize=tiny]{todonotes}

\icmltitlerunning{COMCAT: Towards Efficient Compression and Customization of Attention-Based Vision Models}

\begin{document}

\twocolumn[
\icmltitle{COMCAT: Towards Efficient Compression and Customization of\\ Attention-Based Vision Models}




\begin{icmlauthorlist}
\icmlauthor{Jinqi Xiao}{ru}
\icmlauthor{Miao Yin}{ru}
\icmlauthor{Yu Gong}{ru}
\icmlauthor{Xiao Zang}{ru}
\icmlauthor{Jian Ren}{snap}
\icmlauthor{Bo Yuan}{ru}
\end{icmlauthorlist}

\icmlaffiliation{ru}{Rutgers University}
\icmlaffiliation{snap}{Snap Inc}

\icmlcorrespondingauthor{Jinqi Xiao}{jinqi.xiao@rutgers.edu}

\icmlkeywords{Machine Learning, ICML}

\vskip 0.3in
]



\printAffiliationsAndNotice{}  

\begin{abstract}
Attention-based vision models, such as Vision Transformer (ViT) and its variants, have shown promising performance in various computer vision tasks. However, these emerging architectures suffer from large model sizes and high computational costs, calling for efficient model compression solutions. To date, pruning ViTs has been well studied, while other compression strategies that have been widely applied in CNN compression, \emph{e.g.}, model factorization, is little explored in the context of ViT compression.
This paper explores an efficient method for compressing vision transformers to enrich the toolset for obtaining compact attention-based vision models. Based on the new insight on the multi-head attention layer, we develop a highly efficient ViT compression solution, which outperforms the state-of-the-art pruning methods. For compressing DeiT-small and DeiT-base models on ImageNet, our proposed approach can achieve 0.45\% and
0.76\% higher top-1 accuracy even with fewer parameters. Our finding can also be applied to improve the customization efficiency of text-to-image diffusion models, with much faster training (up to $2.6\times$ speedup) and lower extra storage cost (up to $1927.5\times$ reduction) than the existing works. The code and models are publicly available at  \url{https://github.com/jinqixiao/ComCAT}.

\end{abstract}
\section{Introduction}

Recently the attention-based vision models have achieved comparable or superior performance than the convolution-centered architecture in various computer vision tasks, demonstrating the promising benefit brought by using an attention mechanism~\cite{liu2021video,caron2021emerging,xie2021segformer,carion2020end}. On the downside, these emerging architectures, \emph{e.g.}, vision transformer (ViT) and its variants~\cite{touvron2021training,liu2021swin,li2022efficientformer}, suffer from even larger model sizes and higher computational costs than convolutional neural networks (CNNs), hindering their efficient deployment in many practical resource-constrained scenarios.  

An attractive solution to this challenge is to perform model compression, a strategy that can reduce network size without affecting task performance. Motivated by the huge prior success of compressing CNNs~\cite{hinton2015distilling,han2015deep}, several recent studies~\cite{yin2023gohsp,yu2022unified,hou2022multi} have proposed to apply one (\emph{e.g.}, pruning) or combining several (\emph{e.g.}, pruning and knowledge distillation) compression methods for vision transformers, bringing considerable reduction in model size and/or FLOPs.

Different from the existing works, this paper aims to address the above analyzed efficiency challenge from another perspective -- exploring the low-rankness of the attention-based vision models. To date, a rich set of low-rank compression techniques for CNNs have been proposed in the literature~\cite{kim2015compression,Yin_2021_CVPR,liebenwein2021compressing,yin2022hodec, yin2022batude, xiao2023haloc, xiang2023tdc}. However, consider \ul{1)} there exists a substantial difference on network architecture and operation mechanism, \emph{e.g.}, multi-head attention in ViTs \emph{vs.} channel-wise convolution in CNNs; and \ul{2)} as indicated in \cite{yu2023compressing} and verified by our analysis, many weight matrices in the vision transformers do not exhibit low rankness, it is not clear that whether low-rank compression would bring the satisfied improvement on model efficiency. From the perspective of practical deployment, a question naturally arises: \emph{For compressing attention-based vision models, can exploring model low-rankness provide comparable or even better performance than other methods such as pruning?}

To answer this question and fully unleash the potential of low-rank compression for ViTs and attention-based models, this paper first investigates the low-rankness in the multi-ahead attention layer, and proposes that the head-level low-rankness, instead of weight matrix-level, should be explored. Based on this new insight, we then develop a highly efficient low-rank ViT compression solution with automatic rank selection. Compared with the state-of-the-art ViT pruning methods, the proposed approach can achieve 0.45\% and 0.76\% higher top-1 accuracy even with fewer parameters, for compressing DeiT-small and DeiT-base models on ImageNet dataset, respectively. Furthermore, our finding can also be applied to improve the efficiency for customizing text-to-image diffusion modules~\cite{ruiz2022dreambooth,kumari2022multi}, a recent emerging and important computer vision task, with much faster training (up to $2.6\times$ speedup) and lower extra storage cost (up to $1927.5\times$ reduction) than the state-of-the-art customization solutions.

\section{Related Works}

\textbf{Pruning for Vision Transformers.} To reduce the model size and achieve practical speedup, structured pruning on different substructures of ViT models, \emph{e.g.}, attention heads, blocks, and rows of weight matrices, have been studied in the literature~\cite{hou2022multi,yu2022unified,zhu2021vision,chen2021chasing}. In addition, another research direction proposes to improve model processing speed via using dynamic or static token pruning~\cite{bolya2022token,pan2021scalable,tang2022patch,goyal2020power,pan2021ia}. Recently, Yu \emph{et al.}~\cite{yu2022unified} develop a unified framework to jointly perform pruning, knowledge distillation and block skipping, achieving state-of-the-art ViT compression performance.


\textbf{Low-rank Compression of Weight Matrices ($W$) in NLP Transformers.} Most of the existing studies on low-rank compressed transformers are concentrated in the NLP field. Noach \emph{et al.}~\cite{noach2020compressing} decompose the weight matrices of the pre-trained language models (PLMs) using SVD and perform feature distillation to improve model performance. Ren \emph{et al.}~\cite{ren2022exploring} adopt tensor decomposition to compress PLMs and achieve practical inference speedups. Hsu \emph{et al.}~\cite{hsu2022language} introduce Fisher information to measure the importance of parameters to factorize task-specific PLMs.

\textbf{Low-Rank Approximation for Attention Matrices ($Q$, $K$, $V$).} Another line of works is to perform low-rank approximation for the attention matrices, the intermediate results from the attention mechanism. Different types of approximation schemes, including adding additional projection matrices and sparse approximation, have been investigated~\cite{wang2020linformer, choromanski2020rethinking,chen2021scatterbrain}. As the orthogonal effort from our approach, these methods do not reduce the model sizes of transformers. 

\textbf{Personalized Text-to-Image Diffusion Models.} Recently released text-to-image diffusion models \cite{rombach2022high,ramesh2022hierarchical,saharia2022photorealistic,yu2022scaling} have shown impressive content-generation capability. A very emerging and practical demand is to make these pre-trained models customized for a user-provided specific concept. To that end, 
some efforts leverage transfer learning via fine-tuning all the parameters or introducing a word vector for the new concept~\cite{ruiz2022dreambooth, gal2022image}. However, the large sizes of the diffusion models bring costly training time and high extra storage requirements during the fine-tuning. To improve the efficiency of customization, Kumari \emph{et al.}~\cite{kumari2022multi} propose to only fine-tune the key and value mapping from text to latent features in the cross-attention layers; while freezing other parts. 
\section{Method}
\subsection{Preliminaries}
The attention operation can be viewed as the mapping from a query and a set of key-value pairs to an output. To better extract and learn the information from different representation subspace and spatial regions, the state-of-the-art attention-based vision models adopt multi-head attention (MHA) as:
\begin{equation}
\small
{\rm MHA}(X_Q,X_K,X_V)={\rm Concat}(head_1, \dots, head_h)W^O,
\label{eq:mha}
\end{equation}
where $X_Q,X_K,X_V \in \mathbb{R}^{n \times d_m}$ are the input embedding matrices, $n$ is the sequence length, $d_m$ is the embedding dimension, and $h$ is the number of heads. For each $head_i$, it performs attention operation as follows:
\begin{equation}
\small
\begin{aligned}
head_i&={\rm Attention}( X_QW_{i}^{Q},\ X_KW_{i}^{K},\ X_VW_{i}^{V}) \\
&={\rm Softmax}(\frac{X_QW_{i}^{Q}(X_KW_{i}^{K})^T}{\sqrt{d_k}})X_VW_{i}^{V},
\end{aligned}
\label{eq:self-attention}
\end{equation}
where $W_{i}^{Q}$, $W_{i}^{K} \in \mathbb{R}^{d_m \times d_k}$, $W_{i}^{V} \in \mathbb{R}^{d_m \times d_v}$, $W^{O} \in \mathbb{R}^{hd_v \times d_m}$ are the weight matrices, and $d_k$ and $d_v$ are the dimension of $X_Q$ and $X_K$, respectively. Since $d_k=d_v=\frac{d_m}{h}$, for simplicity we use $d$ instead of $d_k$ and $d_v$. 


\subsection{Exploring Low-Rankness in MHA Layer}
In this subsection, we describe our proposed low-rank MHA for efficient vision models. We first demonstrate the low-rankness of the weight matrices in each attention head, and analyze the limitation when only leveraging the matrix-level low-rankness. Built on these observation and analysis, we then propose to explore the head-level low-rankness and formulate the mechanism. We further detail the procedure of using this finding to improve model efficiency in two important scenarios: compressing vision transformers and customization of diffusion models. 

\begin{figure}[t]
    \centering
    \includegraphics[width=0.95\linewidth]{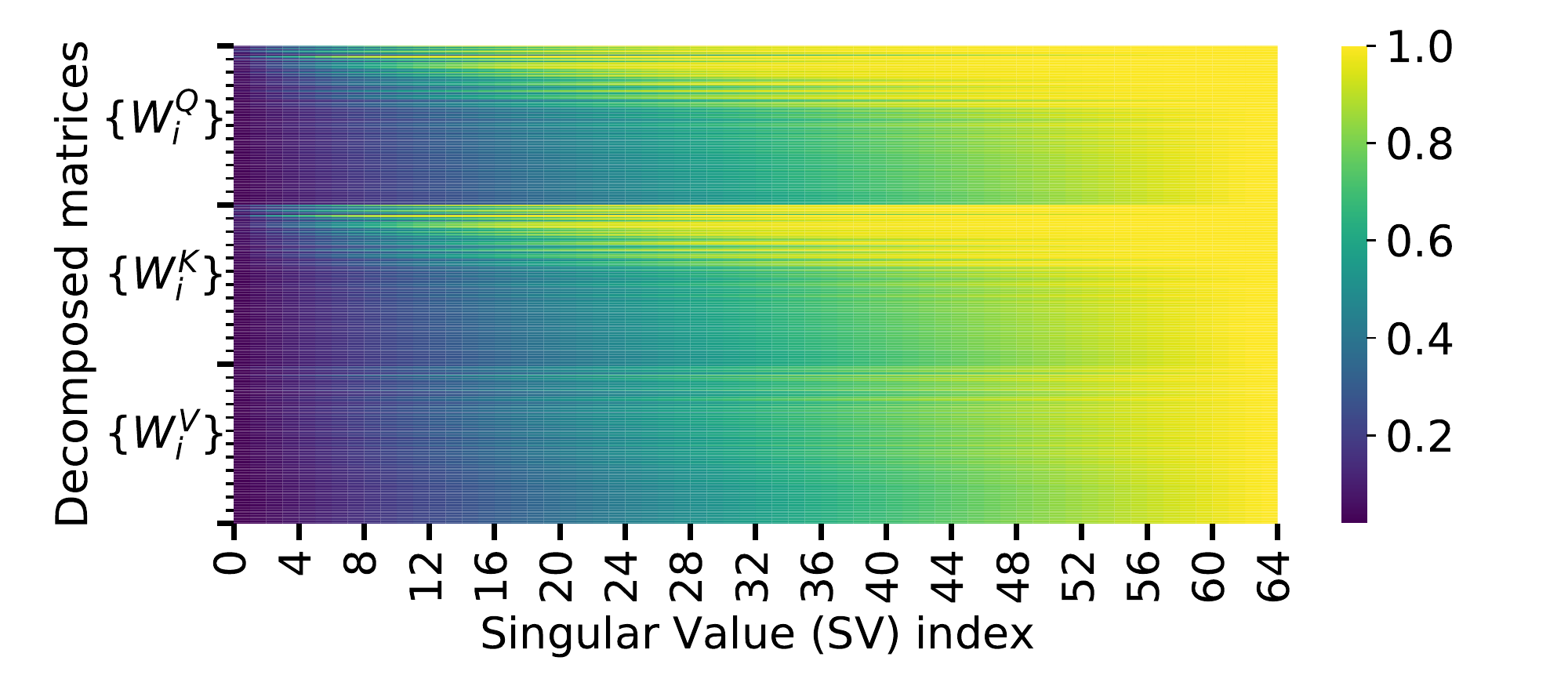}
    \vspace{-4mm}
    \caption{The heatmap of the cumulative singular values after performing SVD for each weight matrices in the MHA layers of the pre-trained DeiT-base model. It is seen that some matrices in some layers exhibit weaker low-rankness than others, implying that directly factorizing individual weight matrix may not be efficient.}
    \label{fig:accu_sv}
\end{figure}

\textbf{Low-Rankness of ${W_{i}^{Q}}$, ${W_{i}^{K}}$, ${W_{i}^{V}}$, ${W^{O}}$.} Low-rankness of the weight matrices has been widely observed in many types of deep learning architectures including CNN and NLP transformers \cite{kim2015compression,ren2022exploring}, inspiring us to explore its potential existence in the attention-based vision models. To that end, we analyze the distributions of the singular values of the weight matrices (${W_{i}^{Q}}$, ${W_{i}^{K}}$, ${W_{i}^{V}}$,${W^{O}}$) in the pre-trained DeiT-base model \cite{touvron2021training}. Figure \ref{fig:accu_sv} shows the heatmap of the cumulative singular values after applying Singular Value Decomposition (SVD) into each weight matrices. It is seen that the phenomenon that most information is concentrated in part of singular values (the largest ones) indeed exist in the weight matrices across different heads and layers, indicating the potential of exploring low rankness of attention-based  models.

\textbf{Limitation of Matrix-Level Low-Rankness.} Based on the above observation, a natural idea is to construct each weight matrix (${W_{i}^{Q}}$, ${W_{i}^{K}}$, ${W_{i}^{V}}$,${W_{i}^{O}}$) in its own low-rank format. However, we argue that the benefit brought this straightforward strategy would not be significant. As shown in Figure \ref{fig:accu_sv}, some types of weight matrices, \emph{e.g.}, $W_{i}^{V}$, do not exhibit sufficient low-rankness, a phenomenon that is also observed in the matrices of the higher layers, thereby limiting the overall potential performance improvement brought by low-rank factorization.

\begin{figure}[t]
    \centering
    \includegraphics[width=0.49\linewidth]{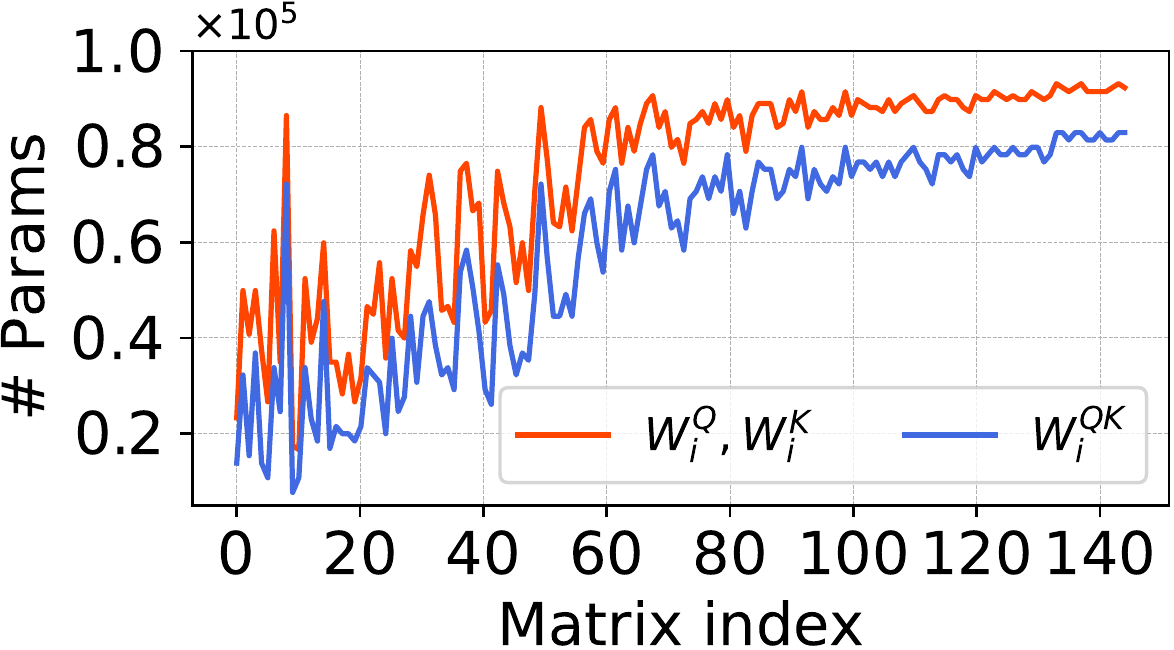}
    \includegraphics[width=0.49\linewidth]{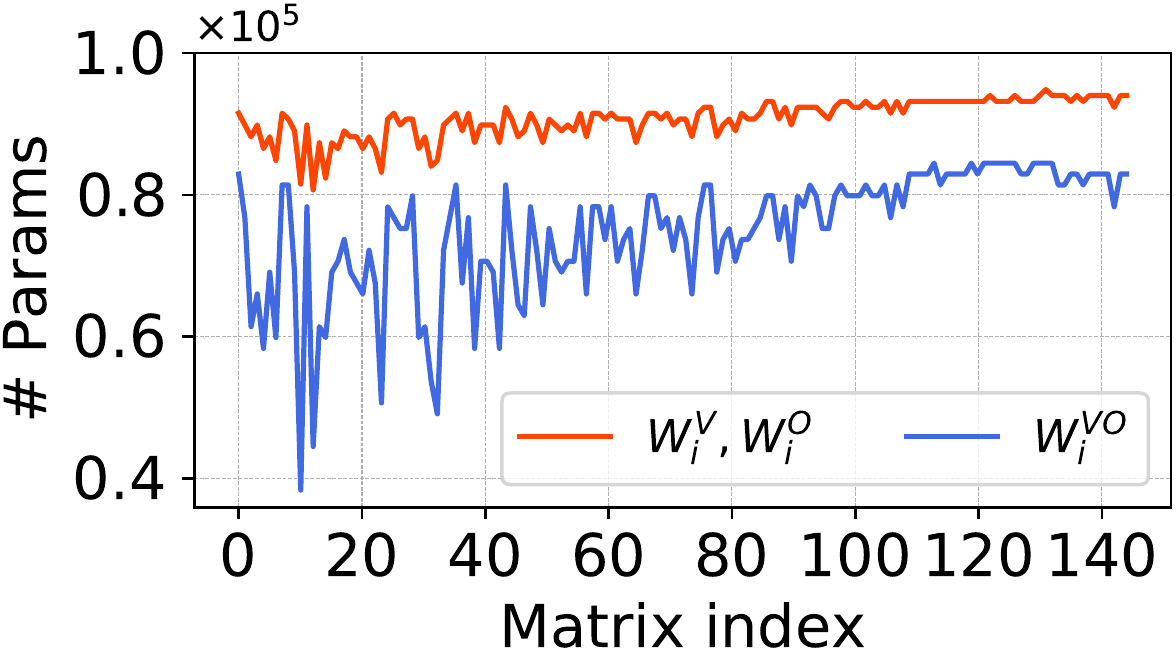}
    \vspace{-3mm}
    \caption{The number of parameters when the ratio of the cumulative singular values reaches 90\% with factorizing $W^Q,W^K,W^V,W^O$, and the corresponding $W^{QK}, W^{VO}$ of the pre-trained DeiT-base. It is seen that exploring head-level low-rankness is more parameter efficient than matrix-level.}
    \label{fig:sv_merged}
    \vspace{-4mm}
\end{figure}

\begin{table}[t]
\vspace{-3mm}
\centering
\setlength\tabcolsep{5pt}
\small
\caption{Top-1 accuracy (\textbf{without fine-tuning}) with factorizing the individual  weight matrices (``Matrix-Level") and the combined matrices (``Head-Level") in the attention layer of the  pre-trained  DeiT-small distilled model (original accuracy is 80.90\%).}
\resizebox{1\linewidth}{!}{
\begin{tabular}{cccccccc}
\toprule
&\multicolumn{2}{c}{\textbf{\# of Params. in MHA $\downarrow$}} & \textbf{20\%}  & \textbf{40\%}  & \textbf{60\%}  & \textbf{80\%} \\
\midrule
&{\multirow{2}{*}{\makecell{\textbf{Top-1(\%)}}}}  &\textbf{Head-Level}  & \textbf{79.81} & \textbf{76.11} & \textbf{63.56} & \textbf{11.5} \\
\cmidrule{3-7}
&& Matrix-Level               & 73.01          & 56.15          & 22.95          & 0.73 \\
\bottomrule
\end{tabular}}
\label{tbl:compr_svd_our}
\vspace{-5mm}
\end{table}

\textbf{Exploring Head-Level Low-Rankness.} To better leverage the low rankness in the MHA layer and fully unleash the potential benefits, we propose to explore the low-rank property at the head level for efficient multi-head attention. Our idea is motivated by the observation that there exists consecutive linear transformations in the attention head, \emph{e.g.}, $X_QW_{i}^{Q}(X_KW_{i}^{K})^T=X_Q(W_{i}^{Q}{W_{i}^{K}}^{T})X_K^{T}$ in Eq. \ref{eq:self-attention}, opening up the opportunities of constructing the combinations of weight matrices, \emph{e.g.}, $W_{i}^{Q}{W_{i}^{K}}^T$, instead of the individual matrix, in the low-rank format. According to linear algebra, such reformulation brings two benefits. \ul{First}, it provides more parameter-efficient low-rank solution. More specifically, for two full-rank matrices $A\in \mathbb{R}^{n\times d}$ and $B\in \mathbb{R}^{d\times n}$, the total number of parameters of their rank-$r$ approximations $A'\in \mathbb{R}^{n\times r}$ and $B'\in \mathbb{R}^{r\times n}$ is $2nr$; while the same rank-$r$ approximation $C'\in \mathbb{R}^{n\times r}$ for $C=AB$ only contains $nr$ parameters. \ul{Second}, it relaxes the constraints of applying low-rankness approximation. Since the low-rankness of $A$ or $B$, instead of both, is sufficient to lead to the low-rank $C$, it indicates that the head-level low-rankness is a more common and feasible opportunity when aiming to leverage low-rankness in the MHA layer. 

\begin{figure*}[t]
\vspace{-3mm}
    \centering
    \includegraphics[width=1\linewidth]{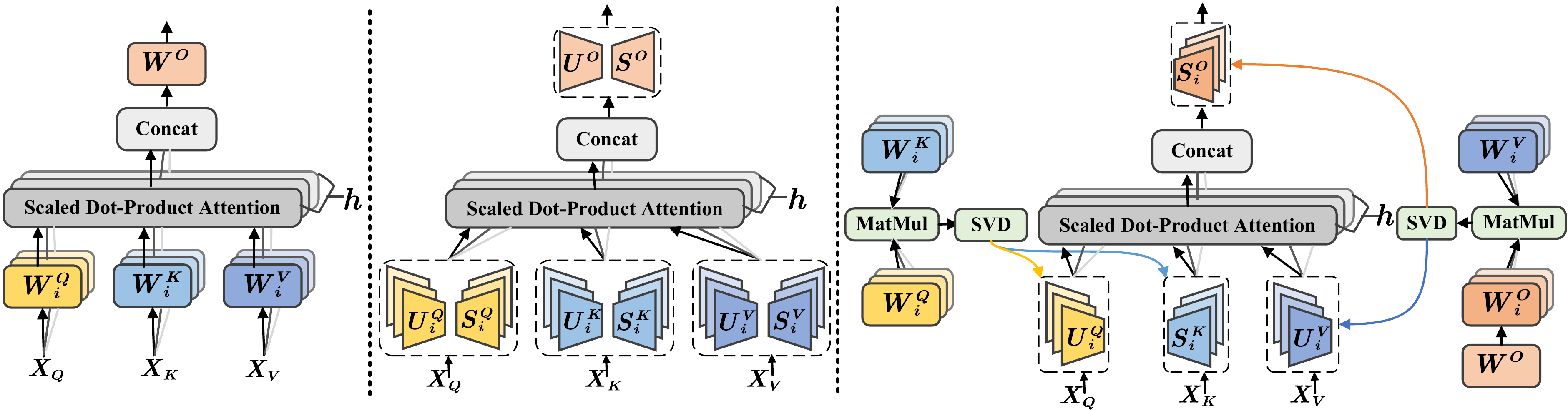}
    \vspace{-9mm}
    \caption{\textbf{(Left)} Standard MHA layer. \textbf{(Middle)} Direct factorization of the individual weight matrices in MHA layer. \textbf{(Right)} Our proposed exploring head-level low-rankness of MHA layer.}
    \label{fig:lowrank}
    \vspace{-4mm}
\end{figure*}

Motivated by these benefits, now we formulate our idea in the context of multi-head attention mechanism. First Eq. \ref{eq:mha} and Eq. \ref{eq:self-attention} can be reformulated as:
\begin{equation}
\small
\begin{aligned}
&{\rm MHA}(X_Q,X_K,X_V)=\sum_{i=1}^{h} head_iW_{i}^{O}\\
&\quad=\sum_{i=1}^{h}{\rm Softmax}(\frac{X_Q(W_{i}^{Q}{W_{i}^{K}}^{T})X_K^T}{\sqrt{d_k}})X_V(W_{i}^{V} W_{i}^{O}),
\end{aligned}
\label{eq:self-attention-new}
\end{equation}
where $W_{i}^{O} \in \mathbb{R}^{d \times d_m}$ and $W^{O}={\rm Concat}(W_{1}^{O},\dots,W_{h}^{O})$.
Recall that a matrix $W \in \mathbb{R}^{in \times out}$ can be low-rank approximated by performing SVD as $W \approx W' = U \Sigma S' = U S$, where $U \in \mathbb{R}^{in \times r}$, $S' \in \mathbb{R}^{r \times out}$, diagonal matrix $\Sigma \in \mathbb{R}^{r \times r}$, $S=\Sigma S' \in \mathbb{R}^{r \times out}$, and $r$ is the rank value. Then the entire multi-head attention (Eq. \ref{eq:self-attention-new}) can be constructed in the low-rank format as:
\begin{equation}
\small
\begin{aligned}
&{\rm MHA}(X_Q,X_K,X_V)\\
& \quad \approx \sum_{i=1}^{h}{\rm Softmax}(\frac{X_Q(U_{i}^{Q}{S_{i}^{K}}^{T})X_K^T}{\sqrt{d_k}})X_V(U_{i}^{V} S_{i}^{O})\\
&\quad={\rm Concat}(head_1', \dots, head_h')W'^O, \quad {\rm where}\\
&head'_i={\rm Attention}( X_QU_{i}^{Q},\ X_KS_{i}^{K},\ X_VU_{i}^{V}) \\
&\quad\quad={\rm Softmax}(\frac{X_QU_{i}^{Q}(X_KS_{i}^{K})^T}{\sqrt{d_k}})X_VU_{i}^{V},\\
&S^O ={\rm Concat}(S_{1}^{O}, \dots, S_{h}^{O}).
\end{aligned}
\label{eq:new_format_svd}
\end{equation}

Here $W_{i}^{Q}{W_{i}^{K}}^{T} \approx U_{i}^{Q}{S_{i}^{K}}^{T}({\rm rank}=r_1)$, $W_{i}^{V} W_{i}^{O} \approx U_{i}^{V} S_{i}^{O}({\rm rank}=r_2)$, $U_{i}^{Q}, S_{i}^{K} \in \mathbb{R}^{d_m \times r_1}$, $U_{i}^{V}, S_{i}^{O} \in \mathbb{R}^{d_m \times r_2}$, and $S^O \in \mathbb{R}^{hr_2 \times d_m}$. 

Notice that as shown in Eq. \ref{eq:new_format_svd}, in addition to exploring the low-rankness of $W_{i}^{QK}=W_{i}^{Q}{W_{i}^{K}}^{T}$, the inter-matrix correlation between $W_{i}^{V}$ and $W_{i}^{O}$ is also considered, bringing the low-rank construction for $W_{i}^{VO}=W_{i}^{V}W_{i}^{O}$. In Figure \ref{fig:sv_merged}, we compare the direct low-rank decomposition of ${W_{i}^{Q}}$ and ${W_{i}^{K}}$ with the decomposition of $W_{i}^{Q}{W_{i}^{K}}^{T}$ and report the required number of parameters after performing low-rank factorization in the case that the ratio of the cumulative singular values reaches $90\%$. By comparing the required number of parameters to reach this threshold, we can evaluate and compare the low-rankness of the original matrices, \emph{i.e.}, fewer parameters indicate better low-rankness. This is because to preserve the same amount of information, \emph{e.g.}, 90\% cumulative singular values, the matrix with better low-rankness requires fewer parameters. As shown in this figure, the number of parameters required for factorizing $W_{i}^{Q}{W_{i}^{K}}^{T}$ (blue line) to reach the $90\%$ threshold is always smaller than that for ${W_{i}^{Q}}$ and ${W_{i}^{K}}$ (red line), indicating that the combination matrix shows better low-rankness. Table \ref{tbl:compr_svd_our} illustrates the benefit of such head-level low-rank MHA mechanism, with its application for fine-tuning-free ViT compression as example. Compared to directly applying SVD to the individual weight matrices, our approach brings much higher model accuracy with the same compression ratio, verifying the two benefits (parameter efficiency and relaxed low-rank constraint) indicated in our prior analysis.


\textbf{Low-Rank MHA for Vision Transformer Compression.} A direct application of our proposed low-rank MHA is to compress vision transformers. In general, for a $b$-block ViT with one MHA layer and one 2-layer feedforward network (FFN) per block, the corresponding compression task using low-rank MHA can be formulated as follows:
\begin{equation}
\small
\begin{aligned}
&\min_{\{W_{i,j}^{QK},W_{i,j}^{VO},W_{k,j}^{FFN}\}_{i=1,j=1,k=1}^{h,b,2}} \mathcal{L}(\{W_{i,j}^{QK},{W_{i,j}^{VO}},W_{k,j}^{FFN}\}) \\ 
&\textrm{s.t.} ~~ \sum_{j=1}^{b}(\sum_{i=1}^{h}\mathcal{C}(\mathcal{R}(W_{i,j}^{QK})) +\mathcal{C}(\mathcal{R}(W_{i,j}^{VO}))\\
& \quad\quad\quad\quad\quad\quad\quad\quad\quad\quad\quad+\sum_{k=1}^{2}\mathcal{C}(\mathcal{R}(W_{k,j}^{FFN}))) \le \varepsilon,
\end{aligned}
\end{equation}
where $\mathcal{L}(\cdot)$, $\mathcal{C}(\cdot)$ and $\mathcal{R}(\cdot)$ are the functions that return the loss, cost (\emph{e.g.}, model size or FLOPs) and rank, 
respectively. $W_{i,j}^{QK}=W_{i,j}^{Q}{W_{i,j}^{K}}^T$, $W_{i,j}^{VO}=W_{i,j}^{V}W_{i,j}^{O}$ and $W_{k,j}^{FFN}$ are the combination matrices of the $i$-th attention head and weight matrix in the FFN in the $j$-th block. It is seen that given the target cost budget ($\varepsilon$) of the compressed ViTs, rank is an important type of hyperparameter that directly determines the accuracy and complexity. In practice, because the huge range of the possible rank values, the proper rank selection for all the blocks and layers of vision transformers is challenging. For instance, there exist $4.53 \times 10^{188}$ rank combinations when performing low-rank compression for DeiT-small model, making manual selection impracticable.

\begin{figure*}
    \centering
    \includegraphics[width=0.9\linewidth]{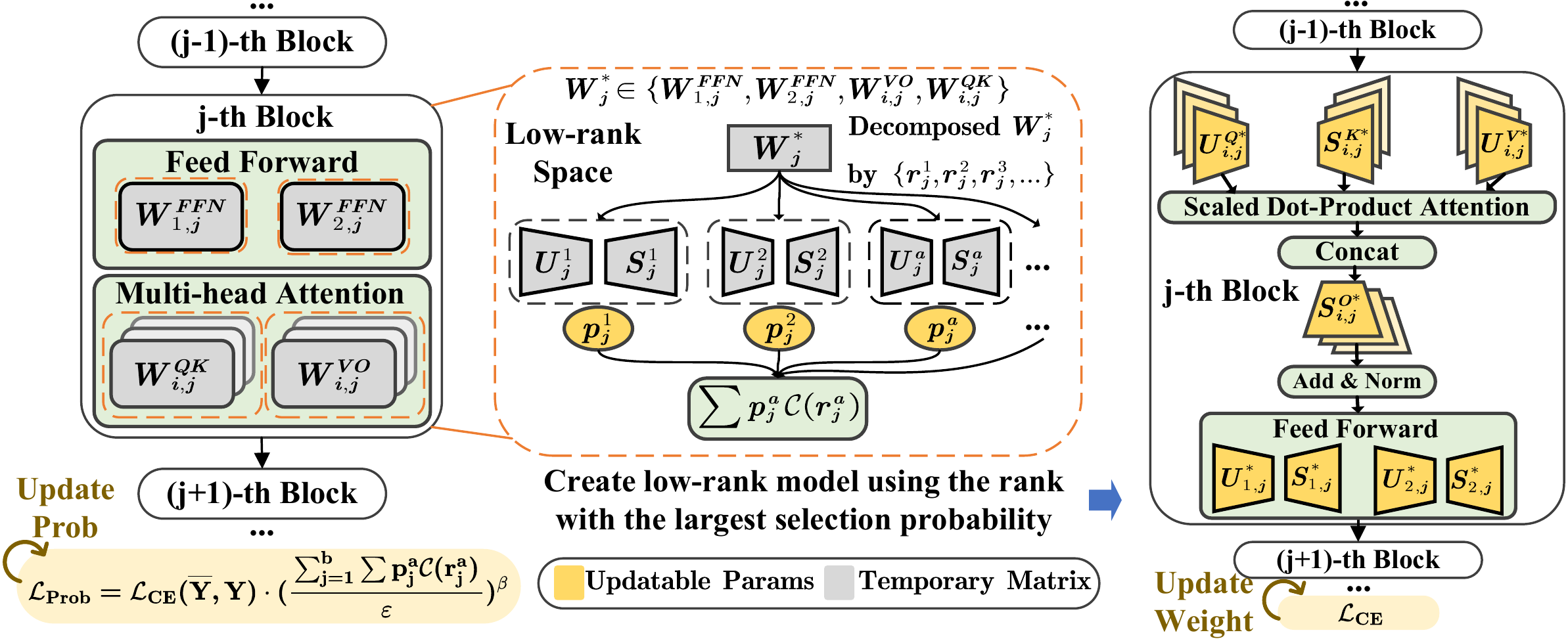}
    \vspace{-4mm}
    \caption{Compressing vision transformer using low-rank MHA layers and automatic rank selection.}
    \label{fig:autorank}
    \vspace{-5mm}
\end{figure*}

To address this challenge, we propose an automatic rank selection approach to efficiently incorporate low-rank MHA into ViT compression. Our key idea is to interpret the automatic rank selection of low-rank compression as a specialized neural architecture search (NAS), considering the fact that the choice of the rank essentially decides the final structure of the compressed ViT. From this insight, the proper rank value can be identified via differentiable sampling-based search, a strategy has been well studied in NAS literature~\cite{liu2018darts,wu2019fbnet,tan2019efficientnet}. 

Figure \ref{fig:autorank} illustrates the overall framework  for automatic rank search for low-rank ViT. Here for simple notation, we use $W_{j}^{*}\in \mathbb{R}^{in_j \times out_j}$ to denote the matrices that need to be decomposed in the $j$-th layer, i.e., $W_{i,j}^{QK}$, $W_{i,j}^{VO}$ and $W_{k,j}^{FFN}$, with the candidate rank set as $R_j = \{r_j^1, r_j^2, r_j^a, \dots\, r_j^{max}\}$. Assume that $r_j^* \in R_j$ is the currently selected rank for $W_j \approx W_j^* = U_j^*S_j^*$ ($U_j^* \in \mathbb{R}^{in_j \times r_j^*}$ and $S_j^* \in \mathbb{R}^{r_j^* \times out_j}$). Then we alternately update the selection probability $P_j=\{p_j^1, p_j^2, p_j^a, \dots\, p_j^{max}\}$ for rank candidates and the parameters of $W_j^*$. To be specific, because $P_j$ is calculated via GumbelSoftmax \cite{jang2016categorical}, \emph{i.e.,} $P_j = {\rm GumbelSoftmax}(\mathbf{\alpha_j})$ with learnable vector $\mathbf{\alpha_j}$, $P_j$ can be updated via minimizing the following loss (with the frozen weight parameters):
\begin{equation}
\small
\begin{aligned}
\mathcal{L}_{Prob} = \mathcal{L}_{CE}(\overline{Y}, Y) \cdot (\frac{\sum_{j=1}^{b}\sum p_j^a\mathcal{C}(r_j^a)}{\varepsilon})^\beta,
\end{aligned}
\label{eq:update_p}
\end{equation}
where $\mathcal{L}_{CE}(\cdot)$ is the cross-entropy loss, $\overline{Y}$ is the final output of the entire model, $Y$ is the ground truth, and $\beta$ is the hyper-parameter controlling overall search process. Here as shown in Figure \ref{fig:autorank}, the calculation of $\overline{Y}=Y_{b}$ is based on considering all the decomposition candidates $(U_{j}^{a}, S_{j}^{a})$ with different rank settings and their selection probabilities. After finishing the probability update, $W_{j}^{*}$ is first factorized via using the rank that corresponds to the largest selection probability, and then updated via minimizing the cross-entropy loss with the frozen rank selection probabilities. The rank settings can be then finally determined after multiple rounds of such alternated update of probabilities and weights.

\textbf{Low-Rank MHA for Personalized Text-to-Image Diffusion.} Our proposed low-rank MHA can also be used for efficiently customizing text-to-image diffusion, an emerging computer vision task that the pre-trained diffusion model can quickly synthesize high-quality visual instantiations of user-defined concepts with few examples of images and guided text prompt. More specifically, given a pre-trained diffusion model $\{W_{diff}\}$ that has been well trained on image set $\{\mathbf{x}\}$ and the condition vector set $\{\mathbf{c}\}$ obtained from text prompt, we aim to minimize the following loss:  
\begin{equation}
\small
 \begin{aligned} 
 &\mathbb{E}_{\epsilon,\mathbf{x_{new}},\mathbf{c_{new}},t} [w_t\| \{W_{new}\}(\alpha_{t}\mathbf{x_{new}}+\sigma_{t}\mathbf{\epsilon},\mathbf{c_{new}})\\
 &\quad\quad\quad\quad\quad\quad\quad\quad\quad\quad\quad\quad\quad\quad\quad-\mathbf{x_{new}} \|_2^2],
 \end{aligned}
 \label{eq:diff-loss}
 \end{equation}
where $\alpha_t$, $\sigma_t$ and $w_t$ are the function of timestep $t$ controlling diffusion process, and $\mathbf{x_{new}}$ and $\mathbf{c_{new}}$ denote the user-provided images and text prompts, respectively, with $|\{\mathbf{x_{new}}\}| \ll |\{\mathbf{x}\}|$ and $|\{\mathbf{c_{new}}\}| \ll |\{\mathbf{c}\}|$. Notice that here $\{W_{new}\}$ is initialized as $f(\{W_{diff}\})$, where $f(\cdot)$ can be either an identity function, meaning that the personalized model $\{W_{new}\}$ is directly initialized as the pre-trained $\{W_{diff}\}$, or a transformation function, indicating that the initialization for $\{W_{new}\}$ is the modification of $\{W_{diff}\}$.   
 


\begin{figure*}
    \centering
    \includegraphics[width=0.95\linewidth]{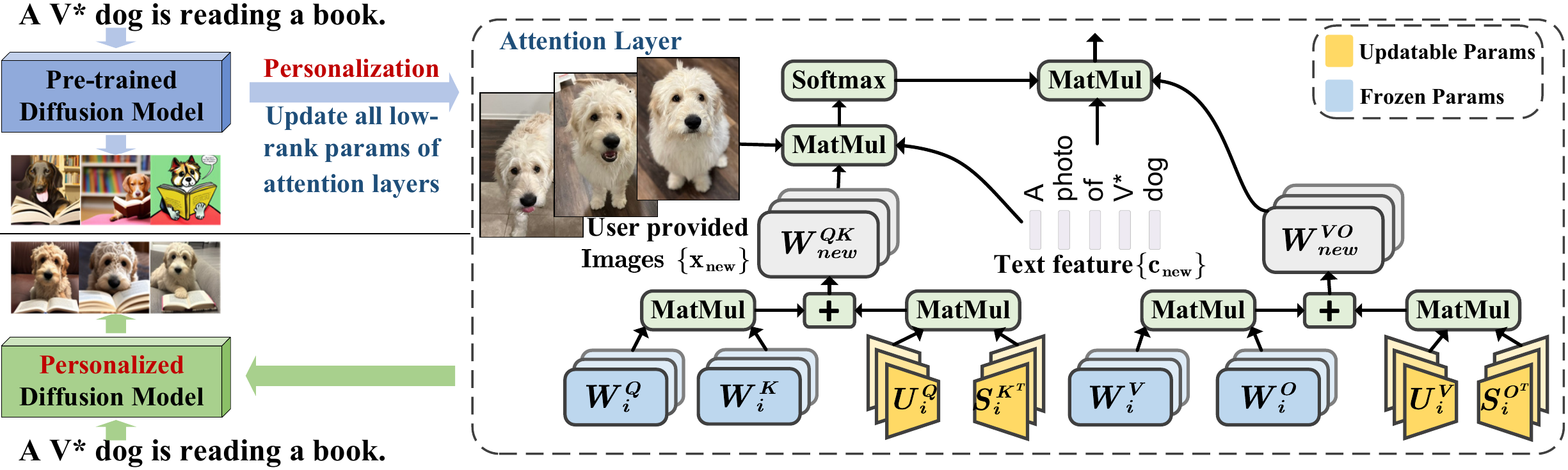}
    \vspace{-2mm}
    \caption{Customizing text-to-image diffusion model using low-rank MHA mechanism. Latent image feature $\mathbf{x_{new}}$ corresponds to $X_Q$ and text feature $\mathbf{c_{new}}$ corresponds to $X_K$ and $X_V$.}
    \label{fig:fintune-diff}
    \vspace{-5mm}
\end{figure*}




As indicated in \cite{kumari2022multi}, updating the entire $\{W_{new}\}$ is very computationally inefficient and easily causes overfitting, due to the large size of $\{W_{diff}\}$ and small size of $\{\mathbf{x_{new}}\}$. Inspired by the insights that \ul{1)} only changing a few parameters is sufficiently to make the diffusion models learn the user-defined concept \cite{kumari2022multi}; and \ul{2)} adding the low-rank component is an efficient fine-tuning strategy for large-size language models (LLMs) in NLP tasks  \cite{aghajanyan2020intrinsic}, we propose to use the low-rank MHA to improve the deployment efficiency of personalized text-to-image diffusion. Figure \ref{fig:fintune-diff} illustrates the overall framework. More specifically, the MHA layer of the personalized model is initialized as:
\begin{equation}
\small
\begin{aligned}
&{\rm MHA}(X_Q,X_K,X_V) = \\
&\quad\quad\sum_{i=1}^{h}{\rm Softmax}(\frac{X_Q(W_{i}^{Q}{W_{i}^{K}}^{T}+U_{i}^{Q}{S_{i}^{K}}^{T})X_K^T}{\sqrt{d_k}})\\
&\quad\quad\quad\quad\quad\quad\quad\quad\quad\quad X_V(W_{i}^{V}W_{i}^{O}+(U_{i}^{V} S_{i}^{O}))
\end{aligned}
\end{equation}
where $\{W_{i}^{Q},W_{i}^{K},W_{i}^{V}, W_{i}^{O}\}$ are obtained from the pre-trained diffusion model $\{W_{diff}\}$, and $\{U_{i}^{Q},S_{i}^{K},U_{i}^{V}, S_{i}^{O}\}$ are the randomly initialized low-rank components. As shown in Figure \ref{fig:fintune-diff}, in the model customization process all the parameters of the pre-trained model $\{W_{diff}\}$ are frozen; while their computation follows the mechanism described in our proposed low-rank MHA. Meanwhile, the added low-rank component $U_{i}^{Q}$, $S_{i}^{K}$, $U_{i}^{V}$ and $S_{i}^{O}$ are updated to make $W_{new}$ adapt for the user-provided new concepts.

\begin{table}
\begin{center}
\vspace{-2mm}
\caption{Comparison between our method and various approaches, including model pruning, sparse training and token reduction, for compressing DeiT-small and DeiT-base on ImageNet. 
}
\vspace{-2mm}
\resizebox{1\linewidth}{!}{
\setlength{\tabcolsep}{1mm}
\begin{tabular}{lcccc}
\toprule
Method & Compression & Top-1 & FLOPs ($\downarrow$\%) & Params ($\downarrow$\%) \\
\toprule
DeiT-small  & Baseline   &79.8    &-   &-     \\
\midrule
\rowcolor{lightgray!30}\textbf{COMCAT (Ours)} & Low-rank & 79.27          & \textbf{51.21} & \textbf{49.98} \\
\rowcolor{lightgray!30}\textbf{COMCAT (Ours)} & Low-rank & 79.58          & 44.93          & 43.82          \\
\rowcolor{lightgray!30}\textbf{COMCAT (Ours)} & Low-rank & \textbf{79.92} & 41.15          & 40.11          \\
UPop~\cite{shi2023upop}          &Pruning          & 	79.6          & 39          & 39                       \\
UVC~\cite{yu2022unified}           &Pruning          & 78.82          & 49.59          & -                        \\
SCOP~\cite{tang2020scop}          &Pruning          & 77.5           & 43.6           & -                         \\
S$^2$ViTE~\cite{chen2021chasing}     &Sparse         & 79.22                &31.63           &33.94\\
ToMe~\cite{bolya2022token}      & Token         & 79.4           & 41.30           &  0                     \\
PS-ViT~\cite{tang2022patch}      & Token         & 79.4           & 43.5           &  0                     \\
HVT~\cite{pan2021scalable}           &Token          & 78.0             & 47.8           &0                        \\
PoWER~\cite{goyal2020power}         &Token          & 78.3             & 41.3           &0                        \\
\hline \hline
DeiT-base  & Baseline   &81.8   &-   &-     \\
\midrule
\rowcolor{lightgray!30}\textbf{COMCAT (Ours)}          & Low-rank & \textbf{82.26} & \textbf{61.68} & \textbf{61.06} \\
CT-GFM ~\cite{yu2023compressing}        &Low-rank          & 81.28          & -          & 40                        \\
MD-ViT~\cite{hou2022multi}        &Pruning          & 81.5          & 60          & -                        \\
UVC~\cite{yu2022unified}                    & Pruning          & 80.57          & 54.5           &-                  \\
VTP~\cite{zhu2021vision}                    &Pruning          & 80.7           & 43.2           &44.44                  \\
S$^2$ViTE~\cite{chen2021chasing}     &Sparse         & 82.22                &33.13           &34.41\\
PS-ViT~\cite{tang2022patch}          &Token          & 81.5          &44.3  & 0                  \\
IA-RED$^2$~\cite{pan2021ia}          &Token          & 80.3           &32.96  & 0                  \\
\bottomrule
\end{tabular}
}
\label{tbl:imagenet}
\vspace{-8mm}
\end{center}
\end{table}
\begin{table*}
\small
\vspace{-2mm}
\caption{Measured speedup for the low-rank compressed DeiT-small and DeiT-base models on different computing platforms.}
\centering
\resizebox{1\linewidth}{!}{
\begin{tabular}{lcccccccc}
\toprule
 {\multirow{2}{*}{{Model}}}  &{\multirow{2}{*}{\makecell{{\#Params} {(M)}}}}  &{\multirow{2}{*}{\makecell{{FLOPs} {(G)}}}}  &{\multirow{2}{*}{\makecell{{Top-1} {(\%)}}}} &\multicolumn{5}{c}{{Throughput (images/s)}}\\ \cmidrule{5-9} &&&&{{Nvidia V100}} &{{Snapdragon 855}} &{{Nvidia JetsonTX2}} &{{ASIC Eyeriss}} &{{FPGA}} \\
 \midrule
 DeiT-small  &21.96    &4.24  &\textbf{79.8}    &974.46     &7.26   &27.37  &24.38   &4.02 \\
 \rowcolor{lightgray!30}\textbf{COMCAT (Ours)}  &\textbf{10.98}  &\textbf{2.07}   &79.27 &\textbf{1512.91} &\textbf{11.00} &\textbf{40.36} &\textbf{39.34} &\textbf{6.66}\\
\toprule
 DeiT-base  &86.38    &16.85  &81.8    &301.34     &1.73  &9.36     &6.14   &0.95\\
 \rowcolor{lightgray!30}\textbf{COMCAT (Ours)}  &\textbf{33.63}   &\textbf{6.46}  &\textbf{82.26}   &\textbf{602.51} &\textbf{4.37} &\textbf{17.80} &\textbf{14.87} &\textbf{2.09}\\
\bottomrule
\end{tabular}
}
\label{tbl:deit-perf}
\vspace{-2mm}
\end{table*}

\begin{figure}[t]
\centering
  \includegraphics[width=0.95\linewidth]{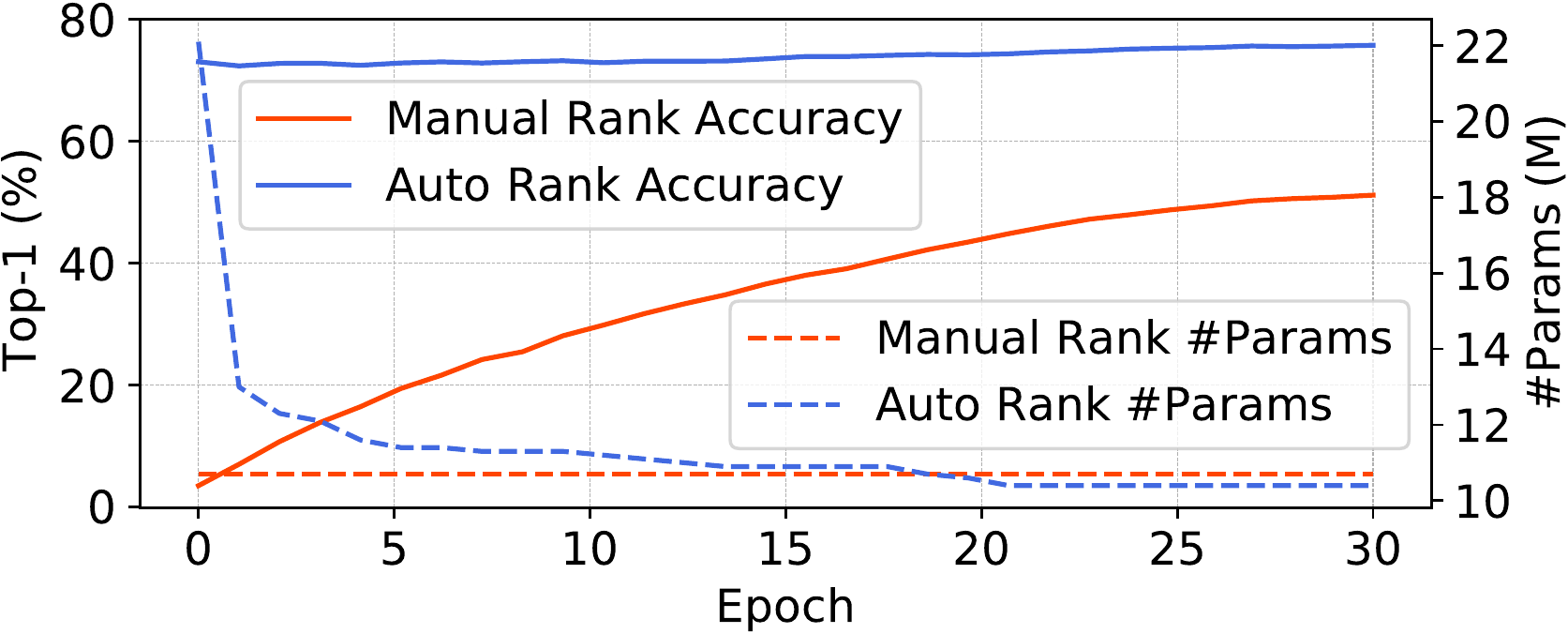}
  \vspace{-4mm}
  \caption{Comparison of the manual rank selection (red lines) and our proposed automatic rank selection method (blue lines) for DeiT-small on the ImageNet-1K dataset. We show the changes of the top-1 accuracy (solid lines) and the number of parameters (\#Params, dashed lines) during training.}
  \label{fig:autovsfixed}
\end{figure}

\begin{figure}[t]
    \includegraphics[width=0.63\linewidth]{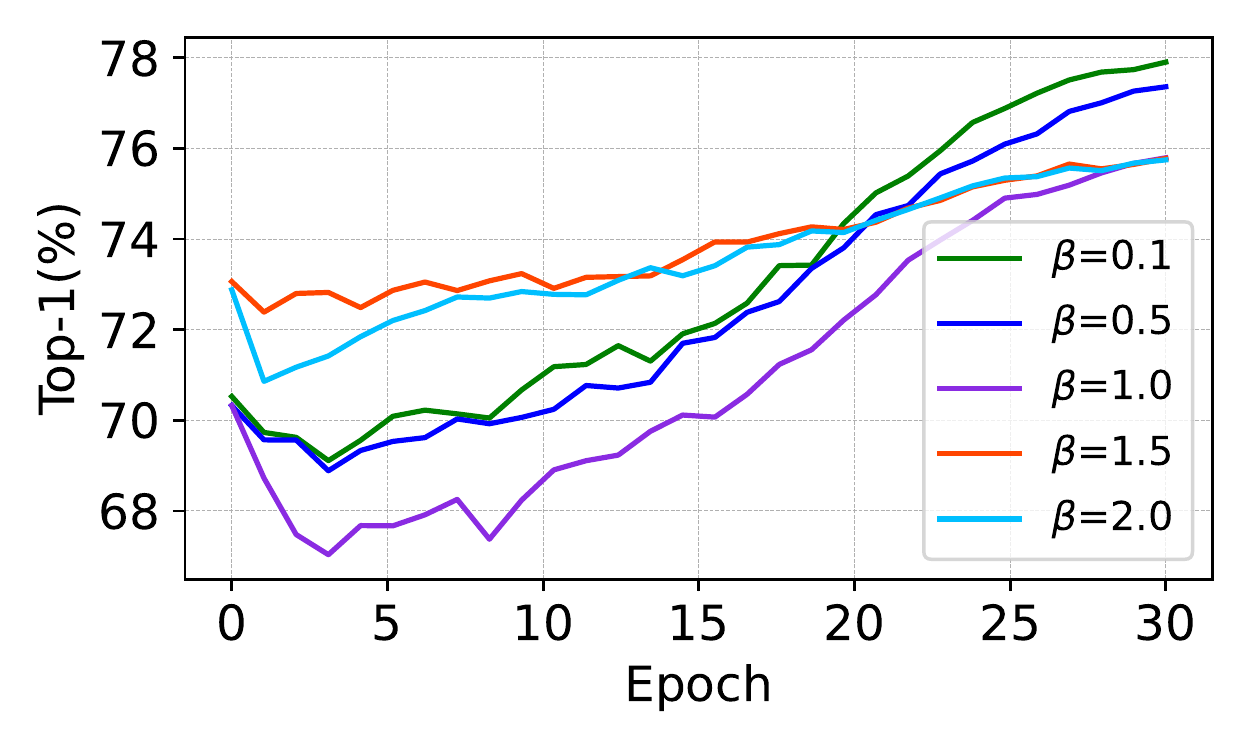}
    \includegraphics[width=0.36\linewidth]{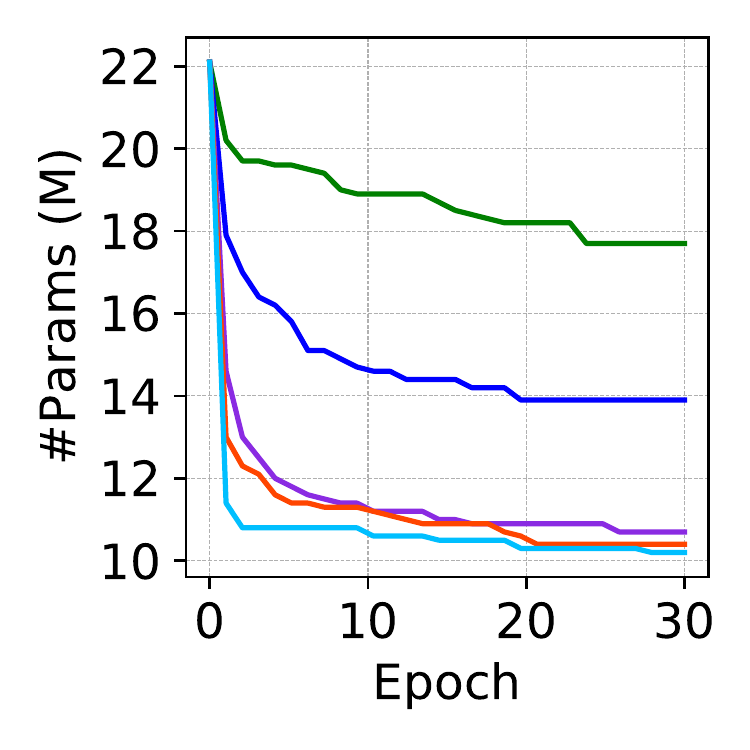}
  \vspace{-8mm}
  \caption{The effect of different beta values on the searching rank of the DeiT-small model.}
  \label{fig:beta}
  \vspace{-4mm}
\end{figure}

\begin{table}
\setlength\tabcolsep{3pt}
\vspace{-3mm}
\caption{Top-1 accuracy and throughput (images/s) on GPU for the compressed DeiT-Small on ImageNet when compressing FFNs using different low-rank methods \textbf{(without using Fine-Tuning)}. It is seen that using SVD to compress FFN layers brings higher accuracy and throughput. Here the throughput is measured on Nvidia V100.}
\resizebox{1\linewidth}{!}{
\begin{tabular}{cccccccc}
\toprule
&\multicolumn{2}{c}{\textbf{\# of Params. in FFN $\downarrow$}}   & \textbf{10\%} & \textbf{20\%} & \textbf{30\%} & \textbf{40\%} & \textbf{50\%} \\ 
\midrule
&{\multirow{2}{*}{\makecell{\textbf{SVD}}}} &\textbf{Top-1 (\%)}  & \textbf{76.75}   & \textbf{73.68}   & \textbf{69.76}   & \textbf{54.63}   & \textbf{21.31}\\ 
\cmidrule{3-8}
&&\textbf{Throughput}    & \textbf{1017.06} & \textbf{1066.81} & \textbf{1129.56} & \textbf{1191.65} & \textbf{1245.19} \\ 
\midrule
&{\multirow{2}{*}{\makecell{\textbf{Tucker}}}} &\textbf{Top-1 (\%)}   & 73.69   & 70.53   & 63.39   & 42.37   & 10.37   \\ 
\cmidrule{3-8}
&&\textbf{Throughput}   & 1017.06 & 1058.94 & 1083.99 & 1141.29 & 1213.87 \\ 
\midrule
&{\multirow{2}{*}{\makecell{\textbf{Tensor Train}}}} &\textbf{Top-1 (\%)}      & 0.81    & 0.53    & 0.35    & 0.16    & 0.14    \\ 
\cmidrule{3-8}
&&\textbf{Throughput}  & 617.47  & 639.70  & 675.39  & 714.79  & 772.87  \\ 
\bottomrule
\end{tabular}}
\label{tbl:compr_ffn}
\end{table}

\begin{figure}[t]
\centering
    \includegraphics[width=\linewidth]{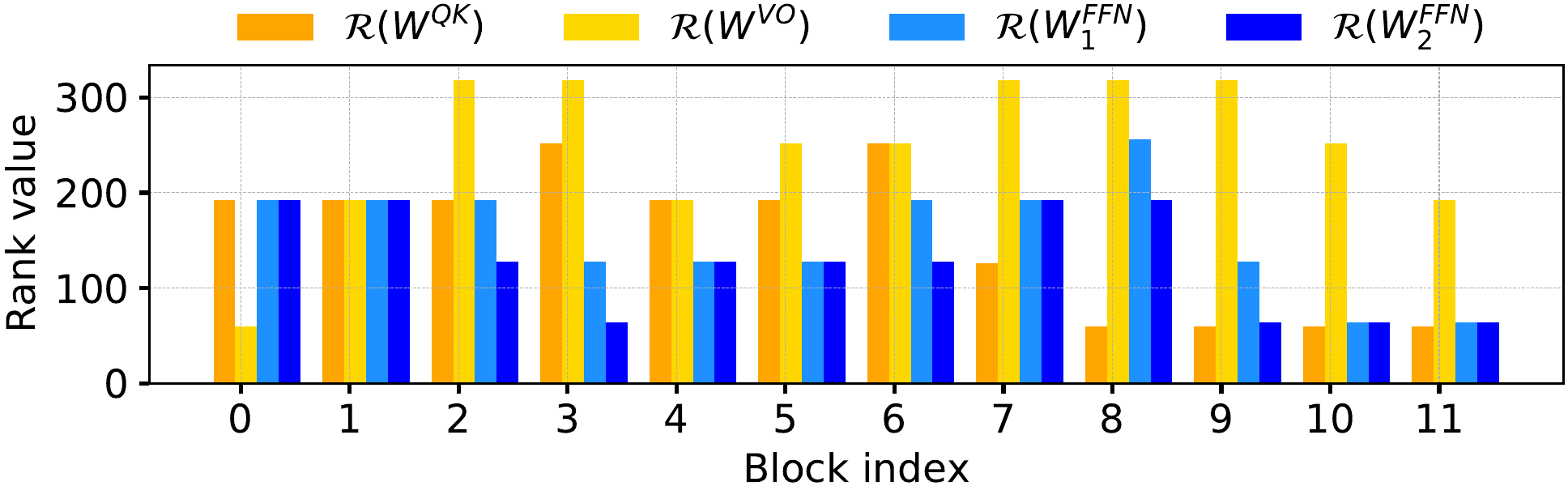}
  \vspace{-7mm}
  \caption{Rank distribution of DeiT-small model.}
   \vspace{-5mm}
  \label{fig:rank-dis-small}
\end{figure}

\begin{figure}[t]
\centering
    \includegraphics[width=\linewidth]{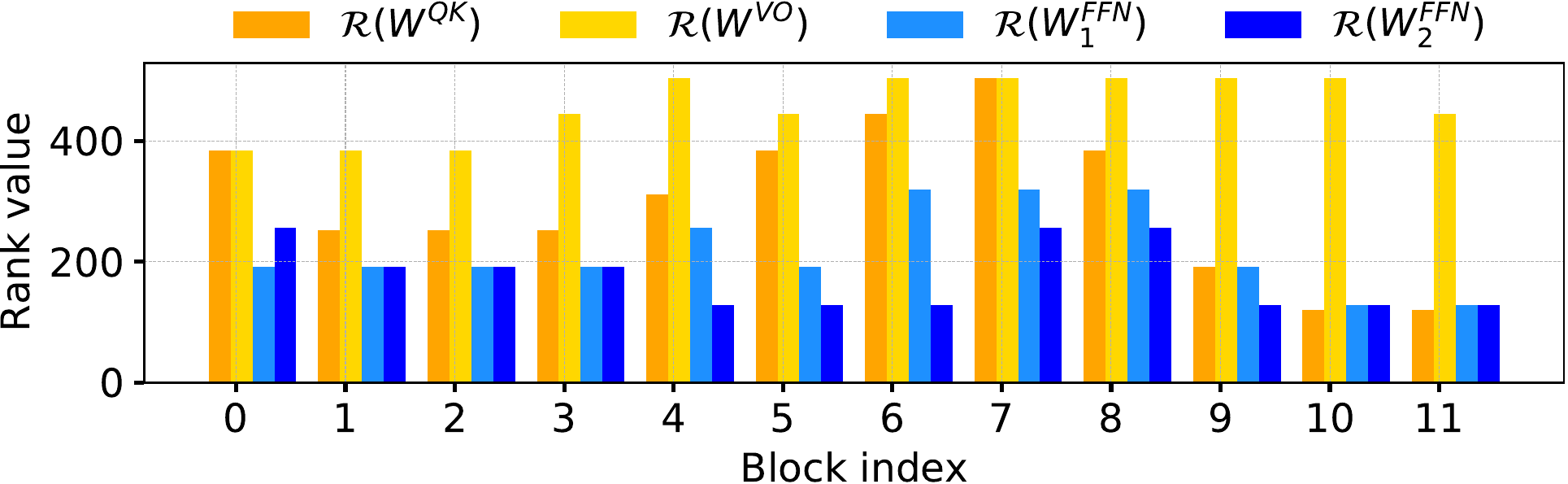}
  \vspace{-8.5mm}
  \caption{Rank distribution of DeiT-base model.}
  \label{fig:rank-dis-base}
  \vspace{-15.5mm}
\end{figure}

\section{Experiments}
\subsection{Image Classification on ImageNet-1K}
\textbf{Setting.}
We first validate our approach on the ImageNet-1K dataset~\cite{deng2009imagenet} for the image classification task. The dataset includes $1.2$M training images and $50$K validation samples. We adopt the baseline models, \emph{i.e.,} dense networks without compression, and training recipe from DeiT~\cite{touvron2021training} since they show promising results on training the transformer models by only using the ImageNet-1K without other large-scale datasets for pre-training and thus are widely adopted.  
We compare our approach with previous state-of-the-art ViT compression methods, including low-rank~\cite{yu2023compressing}, model pruning~\cite{hou2022multi,yu2022unified,zhu2021vision,tang2020scop}, sparse training \cite{chen2021chasing} and token pruning~\cite{bolya2022token,pan2021scalable,tang2022patch,goyal2020power,pan2021ia}.

\textbf{Implementation Details.} The whole process of our method consists of two steps: searching ranks and fine-tuning. We conduct the automatic rank selection algorithm  to the pre-trained DeiT model to produce the low-rank model under the given constraint. In the fine-tuning process, the initial learning rate is set as $0.0001$ and decreases to the minimum learning rate of $0.000001$ with the Cosine scheduler. The weight decay for training the compressed DeiT-small is set as $0.005$. The rest of the training hyper-parameters are consistent with DeiT~\cite{touvron2021training}.

\textbf{Comparison Results.} Table~\ref{tbl:imagenet} shows the performance of different compression methods on the ImageNet dataset. Compared with the previous state-of-the-art automatic pruning method UVC~\cite{yu2022unified}, our compressed models enjoy $0.45\%$ and $1.69\%$ higher top-1 accuracy with much larger FLOPs reduction on DeiT-small and DeiT-base, respectively. Additionally, our approach can significantly reduce the model parameters, \emph{i.e.,} $49.98\%$ reduction for DeiT-small and $61.06\%$ for DeiT-base, while the compressed model from UVC~\cite{yu2022unified} can not.
Compared with low-rank work of CT-GFM~\cite{yu2023compressing}, our method achieves 0.98\% accuracy increase with much fewer number of parameters.
Compared with sparse training work of S$^2$ViTE~\cite{chen2021chasing}, with similar top-1 accuracy, we achieve much larger FLOPs reduction as $19.58\%$ and $28.55\%$ and parameters reduction as $16.04\%$ and $26.65\%$ on DeiT-small and DeiT-base, respectively. Compared with the work of PS-ViT \cite{tang2022patch} for token reduction, our method also achieves higher top-1 accuracy as $0.18\%$ and $0.76\%$ with much fewer FLOPs and model parameters on DeiT-small and DeiT-base, respectively.

\textbf{Practical Speedups on Various Hardware Platforms.}
We further measure the practical speedups of our compressed models on various computing hardware platforms, including Nvidia Tesla V100, Nvidia Jetson TX2, Android mobile phone (Snapdragon 855, 4 Cortex-A76 + 4 Cortex-A55), ASIC accelerator Eyeriss \cite{chen2016eyeriss}, and FPGA (PYNQ Z1) in Table \ref{tbl:deit-perf}. Here the performance of Eyeriss is reported via using Timeloop \cite{parashar2019timeloop} with 45nm CMOS technology setting. Our compressed DeiT-small and DeiT-base models achieve significant speedups across different platforms. For example, on Snapdragon 855, our compressed DeiT-base obtains $2.52\times$ speedup than the baseline model with even higher top-1 accuracy on ImageNet. Such results demonstrate the practical effectiveness of our low-rank compression solution.

\subsection{Ablation Analysis for ViT Low-Rank Compression}
\textbf{Automatic Rank Selection \emph{vs.} Manual Rank Selection.} To demonstrate the superiority of our automated rank selection method, we compare it with the fixed rank method. Figure~\ref{fig:autovsfixed} shows the variation curves of the top-1 accuracy on ImageNet-1K and the number of parameters for both methods during training. Our method has a smaller loss of accuracy, and the number of parameters of the model gradually converges to the target value (10.5M). In contrast, the model based on fixed-rank decomposition loses more accuracy from the beginning, resulting in poor performance. Therefore, we can conclude that our method can search for a better rank combination under the constraints.

\textbf{Hyper-Parameter for Searching Ranks.} We also explore the effect of hyper-parameter $\beta$ mentioned by Eq. \ref{eq:update_p} on searching rank process. Figure~\ref{fig:beta} shows the convergence of the searching process with respect to different $\beta$. It can be seen that when $\beta \ge 1$, the number of parameters of the model can converge to the target value quickly, and the final accuracy of the models is basically the same. However, when $\beta=1.5$, the accuracy curve of the model is relatively smooth, therefore, we think 1.5 is a relatively better value for $\beta$. The final rank distribution of DeiT-small and DeiT-base are shown in Figure~\ref{fig:rank-dis-small} and Figure~\ref{fig:rank-dis-base}.

\textbf{SVD \emph{vs.} Higher-order Tensor Decomposition.} For ViT compression, in addition to the MHA, we apply the low-rank compression to the FFN (Feed-Forward Network), and the rank selection for FFN is also included in our proposed automatic rank determination mechanism. In order to find the optimal low-rank decomposition method from various low-rank decomposition methods such as SVD, Tucker decomposition, Tensor Train decomposition, we evaluate the Top-1 accuracy and throughput on GPU for the compressed DeiT-Small on ImageNet dataset when compressing FFNs using different low-rank methods. As shown in Table \ref{tbl:compr_ffn}, with the same compression rate, using SVD brings higher accuracy (without fine-tuning) than using Tucker decomposition and Tensor Train decomposition with better throughput on GPU.

\begin{table*}[t]
\small
\begin{center}
\caption{Comparison of the training cost, \emph{i.e.,} training time, GPU memory, and extra storage for each concept, and FID~\cite{parmar2021buggy} for various methods.
Given a few images of a new concept, we generate images corresponding to the text prompt, \emph{e.g.,} A photo of V * dog.
The number in $(\cdot)$ denotes the number of images involved in training. ``V*" is a unique identifier followed by the class name of the subject that is used to identify the object to be learned. Except for Goldendoodle, the rest of the images are from CustomDiffusion.} 
\resizebox{1\linewidth}{!}{
\setlength{\tabcolsep}{1mm}
\begin{tabular}{lccccccccccccc}
\toprule
& \multicolumn{3}{c}{{Training Cost}} & &\multicolumn{8}{c}{{FID}}\\ 
 \cmidrule{2-4} \cmidrule{6-13}
{Method} &{\multirow{2}{*}{\makecell{{Training}\\{Time (s)}}}} &{\multirow{2}{*}{\makecell{{GPU}\\{Memory (MB)}}}} & {\multirow{2}{*}{\makecell{{Extra}\\{Storage (MB)$^*$}}}} & &{\multirow{2}{*}{\makecell{{Teddy}\\{Bear (7)}}}} &{\multirow{2}{*}{\makecell{{Tortoise}\\{Plushy (12)}}}} &{\multirow{2}{*}{\makecell{{Wooden}\\{Pot (4)}}}}  &{\multirow{2}{*}{\makecell{{Barn}\\{ (7)}}}} &{\multirow{2}{*}{\makecell{{Cat}\\{ (5)}}}} &{\multirow{2}{*}{\makecell{{Dog}\\{ (10)}}}} &{\multirow{2}{*}{\makecell{{Golden-}\\{doodle (4)}}}} &{\multirow{2}{*}{\makecell{{Average}}}}  \\\\
\midrule
\rowcolor{lightgray!30}\textbf{COMCAT (Ours)} 	&\textbf{193} &\textbf{11765}         &\textbf{6} & &76.61	&\textbf{168.42}	&91.13  &\textbf{42.5}	&139.51	&\textbf{86.68}	&\textbf{119.49}  &\textbf{101.23}\\
CustomDiffusion  &237          &11807       & 75 &&93.92	&196.15	&136.76  &52.23	&\textbf{127.93}	&146.14	&150.28  &128.06\\
DreamBooth  &502          &30979            &11565 &&\textbf{62.25}	&186.52	&\textbf{89.31}  &47.93	&150.79	&89.31 &179.33  &118.03\\
\bottomrule
\end{tabular}}
\label{tbl:diff-FID}
\end{center}
\vspace{-2mm}
\scriptsize{*The extra storage means that, given a pre-trained model, the extra storage requirement when the pre-trained model is further fine-tuned to adapt to a new customized concept. Notice that here we follow the same definition for extra storage cost used in Customize Diffusion \cite{kumari2022multi}. That is, in customization scenario, because the pre-trained model needs to be always preserved for future more new concepts, any modification on the pre-trained model for the current new concept is viewed as extra storage cost. The amount of extra storage is obtained via direct measurement of file size.}
\end{table*}

\begin{table}[t]
\setlength\tabcolsep{3pt}
\vspace{-2mm}
\centering
\caption{MS-COCO FID evaluation with fine-tuned models is a standard evaluation metric for text-to-image models. The pre-trained diffusion model has FID as $32.9$ with the same settings of $50$ PNDM \cite{liu2022pseudo} sampling steps and scale as $6$. }
\resizebox{1\linewidth}{!}{
\begin{tabular}{ccccccccc}
\toprule
  & {Pre-trained}  & {Goldendoodle} & {Barn} & {Cat}  \\
\midrule
{MS-COCO FID} &32.90 & 32.84                 & 31.88         & 30.98       \\
\toprule
& {Dog} & {TeddyBear} & {TortoisePlushy} & {WoodenPot}\\
\midrule
{MS-COCO FID} & 30.85        & 30.89              & 32.30                      & 31.29   \\
\bottomrule
\end{tabular}
}
\label{tbl:diff-FID-mscoco}
\vspace{-5mm}
\end{table}

\subsection{Personalized Text-to-Image Diffusion Models}
\begin{figure*}[ht]
\small
\centering
\setlength\tabcolsep{1.5pt}
\begin{tabular}{ccccccccc}
\toprule
&{Target Images} &&{COMCAT (Ours)} &&{Custom Diffusion} &&{DreamBooth}\\
\midrule
&\multirow{6}{*}{
\begin{minipage}{0.26\columnwidth}
    \vspace{+6mm}
    \includegraphics[width=\linewidth]{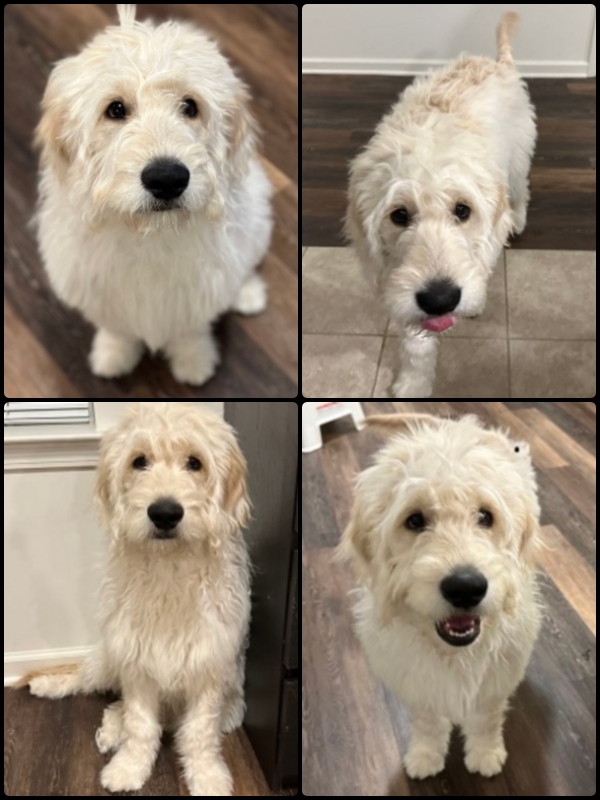}
\end{minipage}}
&&
\begin{minipage}{0.55\columnwidth}
    \includegraphics[width=\linewidth]{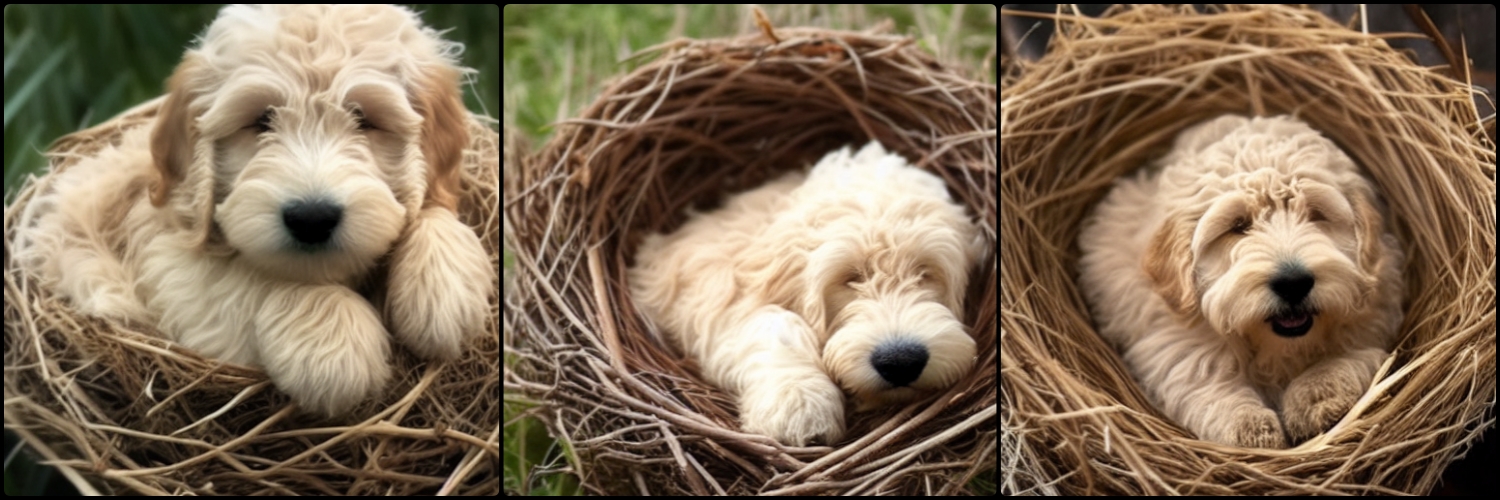}
\end{minipage}
&&
\begin{minipage}{0.55\columnwidth}
    \includegraphics[width=\linewidth]{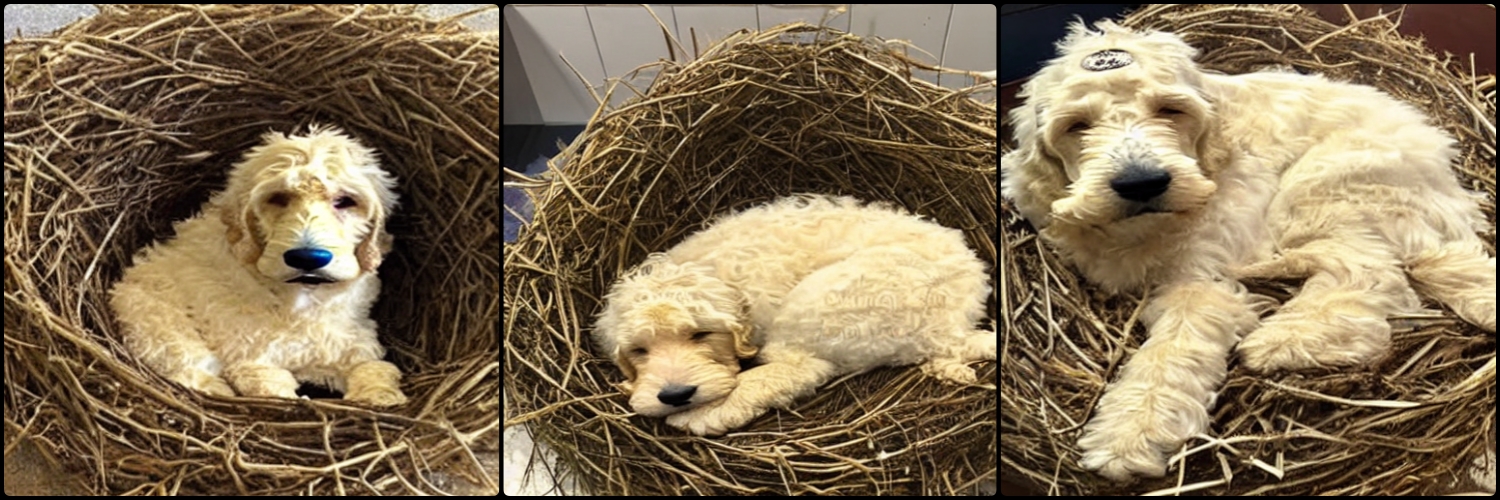}
\end{minipage}
&&
\begin{minipage}{0.55\columnwidth}
    \includegraphics[width=\linewidth]{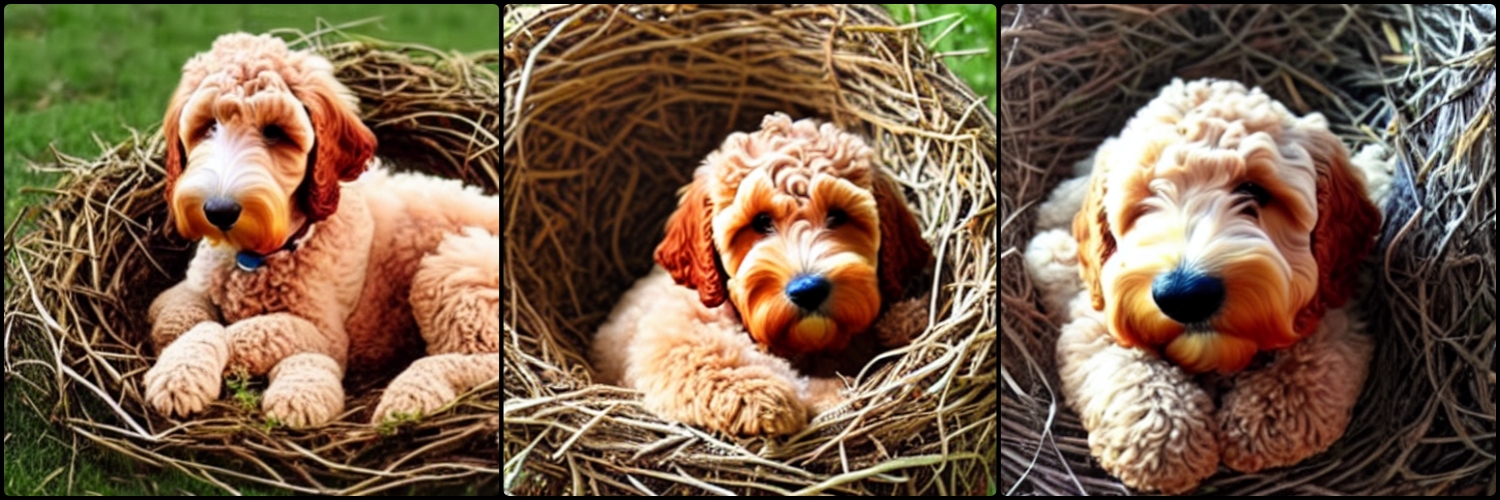}
\end{minipage}\\

&&&\multicolumn{5}{c}{\textbf{V* Goldendoodle sleeping in a bird nest}}\\
\cmidrule{4-8}
&&&
\begin{minipage}{0.55\columnwidth}
    \includegraphics[width=\linewidth]{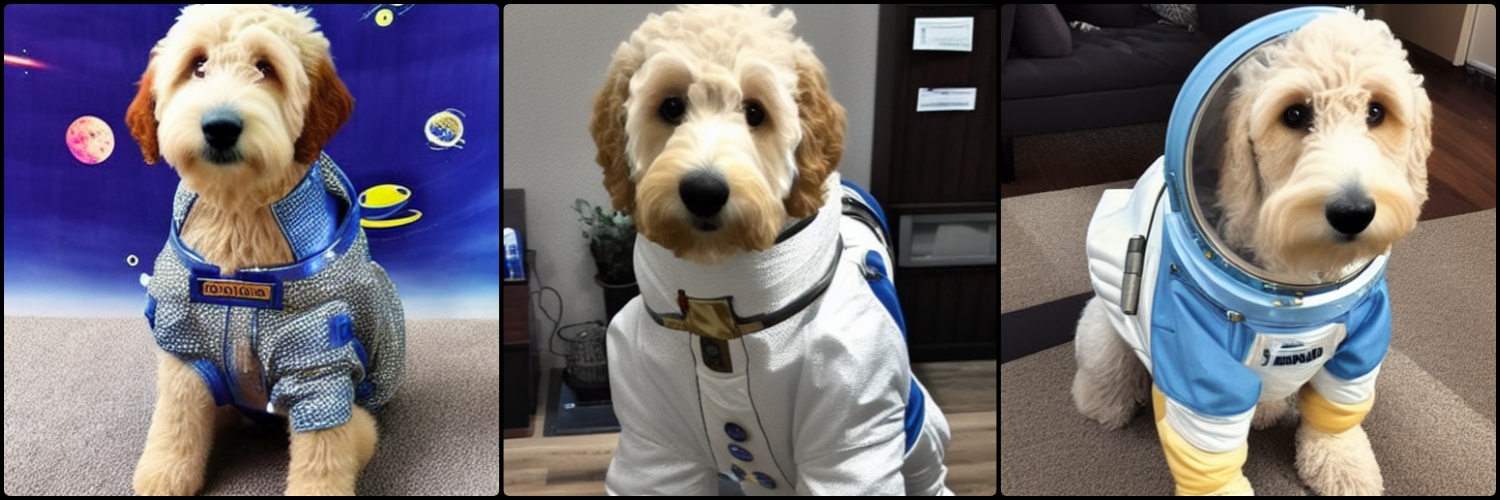}
\end{minipage}
&&
\begin{minipage}{0.55\columnwidth}
    \includegraphics[width=\linewidth]{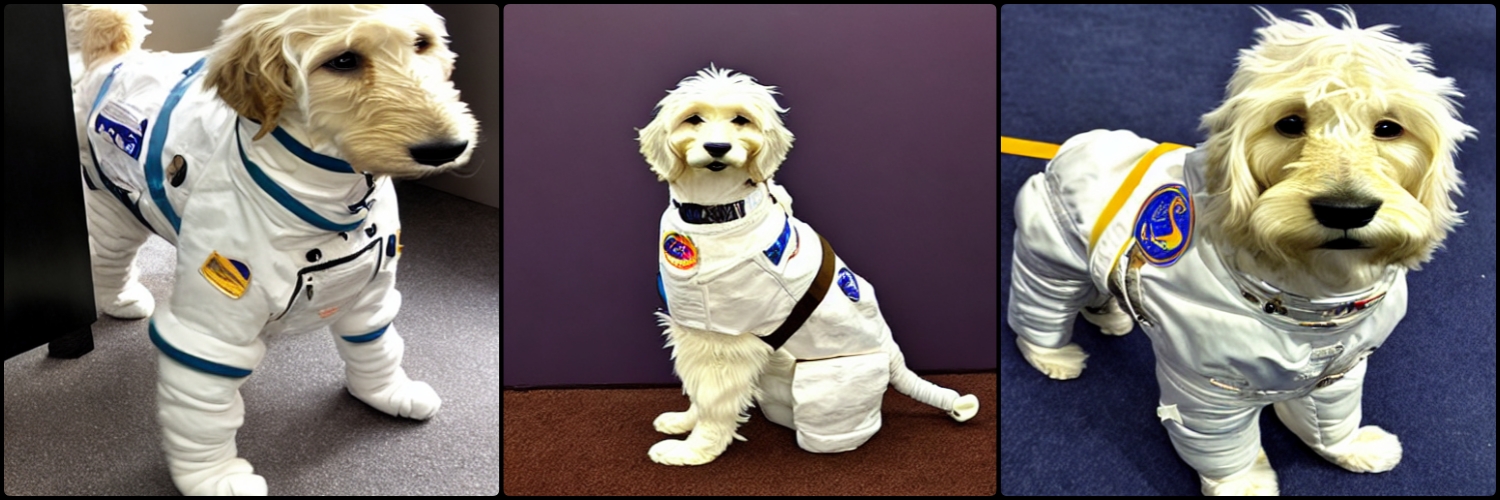}
\end{minipage}
&&
\begin{minipage}{0.55\columnwidth}
    \includegraphics[width=\linewidth]{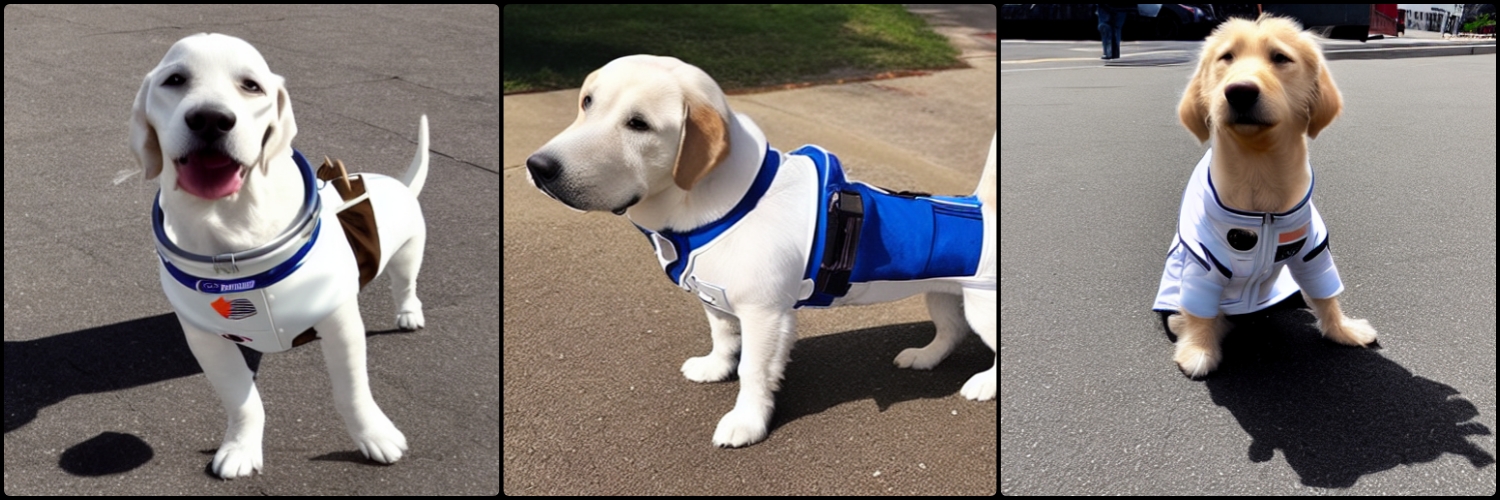}
\end{minipage}\\
&&&\multicolumn{5}{c}{\textbf{V* Goldendoodle wearing space suit}}\\
\cmidrule{4-8}
&&&
\begin{minipage}{0.55\columnwidth}
    \includegraphics[width=\linewidth]{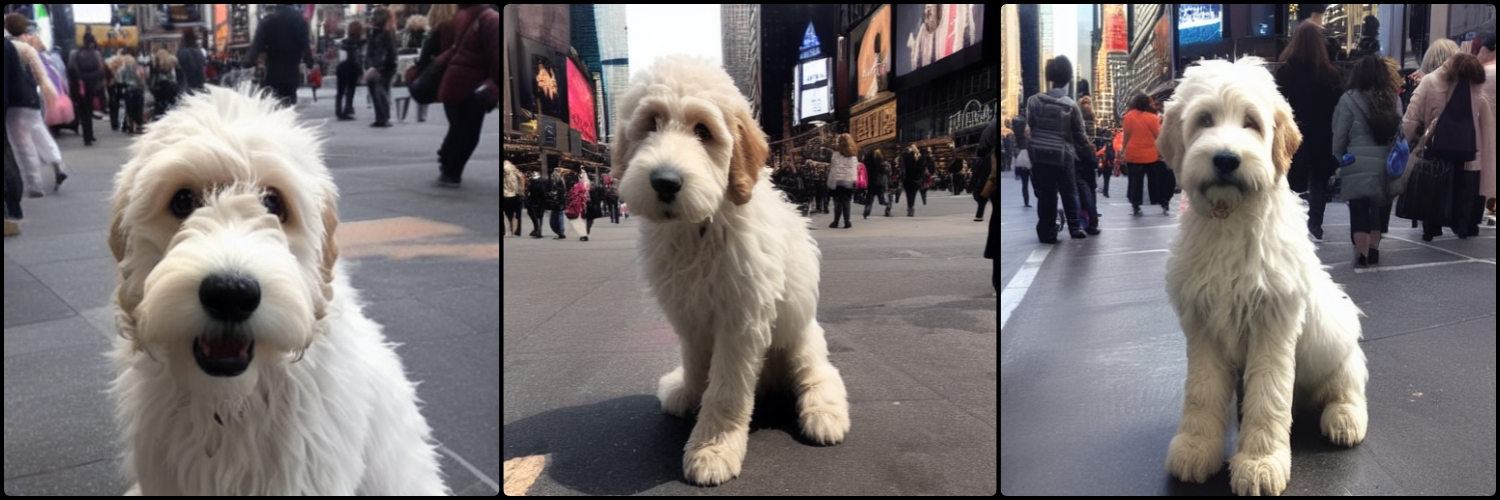}
\end{minipage}
&&
\begin{minipage}{0.55\columnwidth}
    \includegraphics[width=\linewidth]{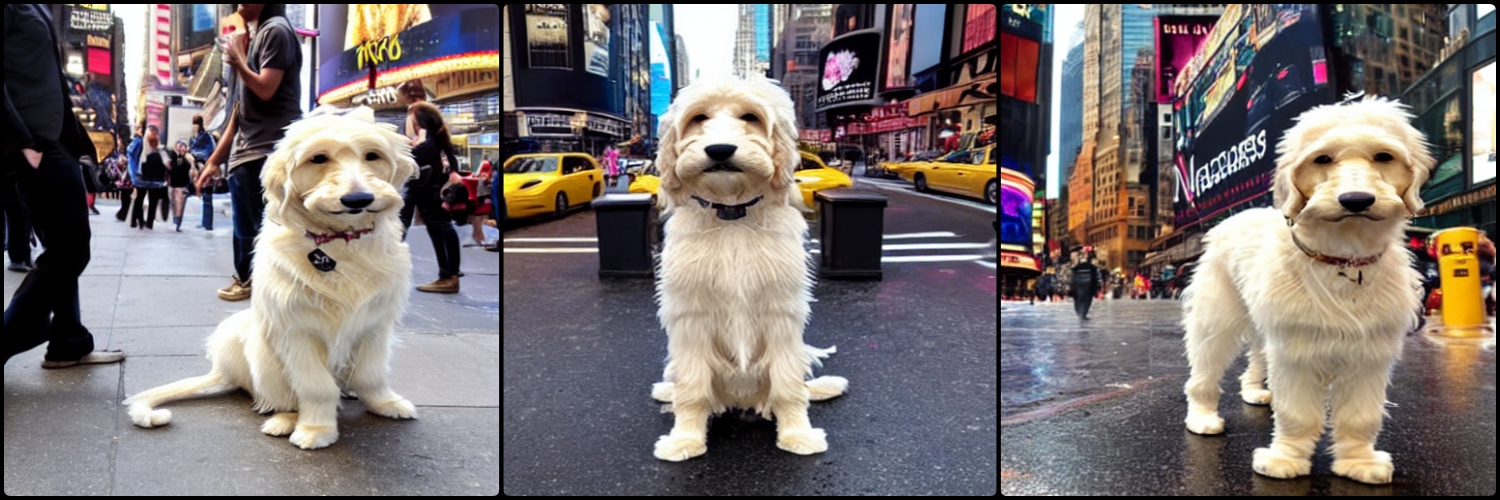}
\end{minipage}
&&
\begin{minipage}{0.55\columnwidth}
    \includegraphics[width=\linewidth]{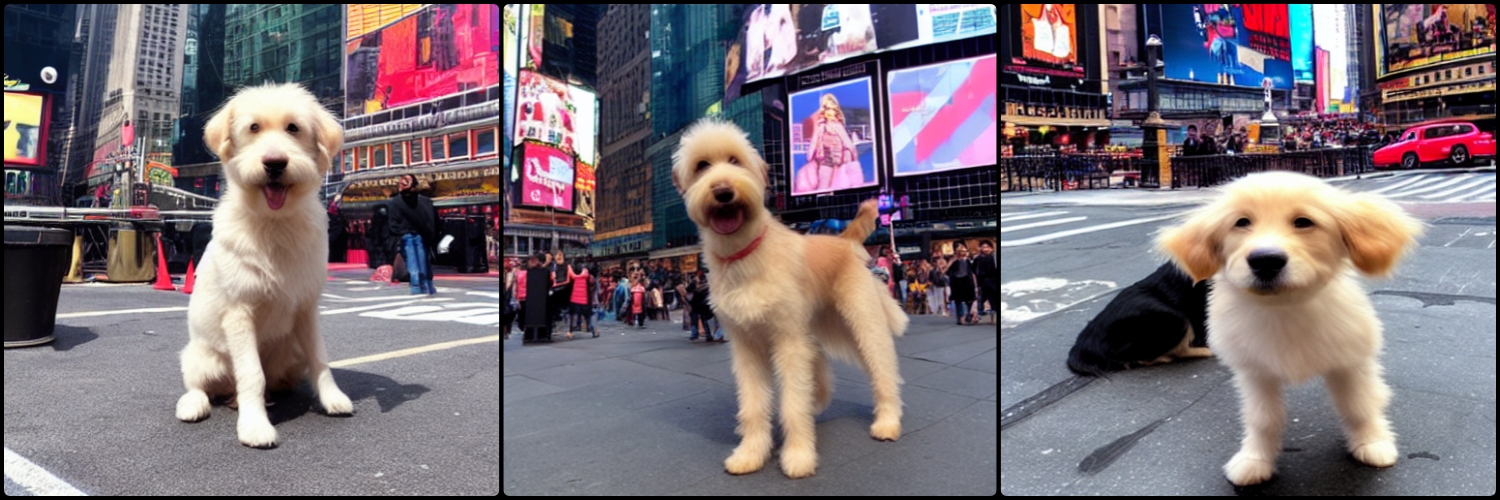}
\end{minipage}\\
&&&\multicolumn{5}{c}{\textbf{V* Goldendoodle in Times squares}}\\
\bottomrule
\end{tabular}
\caption{Fine-tuning results. All methods were trained on 1 A6000 GPU for 500 steps, and the training consumption is shown in Table \ref{tbl:diff-FID}. All images were generated with 50 steps of PNDM \cite{liu2022pseudo} sampler and guidance scale is 7.}
\label{fig:finetune-diff-gd}
\end{figure*}
\begin{figure*}[t]
\centering
\setlength\tabcolsep{1.5pt}
\begin{tabular}{cccccccccccc}
\toprule
&{Training Cost} &&\textbf{3 minutes} (1 A6000 GPU) &&6 minutes (2 A100 GPUs) &&1 hour (4 A100 GPUs)\\
\cmidrule{1-2}\cmidrule{4-4}\cmidrule{6-6}\cmidrule{8-8}
&{Target Images} &&{COMCAT (Ours)} &&{Custom Diffusion$^\dag$} &&{DreamBooth$^\dag$} \\
\midrule
&
\begin{minipage}{0.47\columnwidth}
    \includegraphics[width=\linewidth]{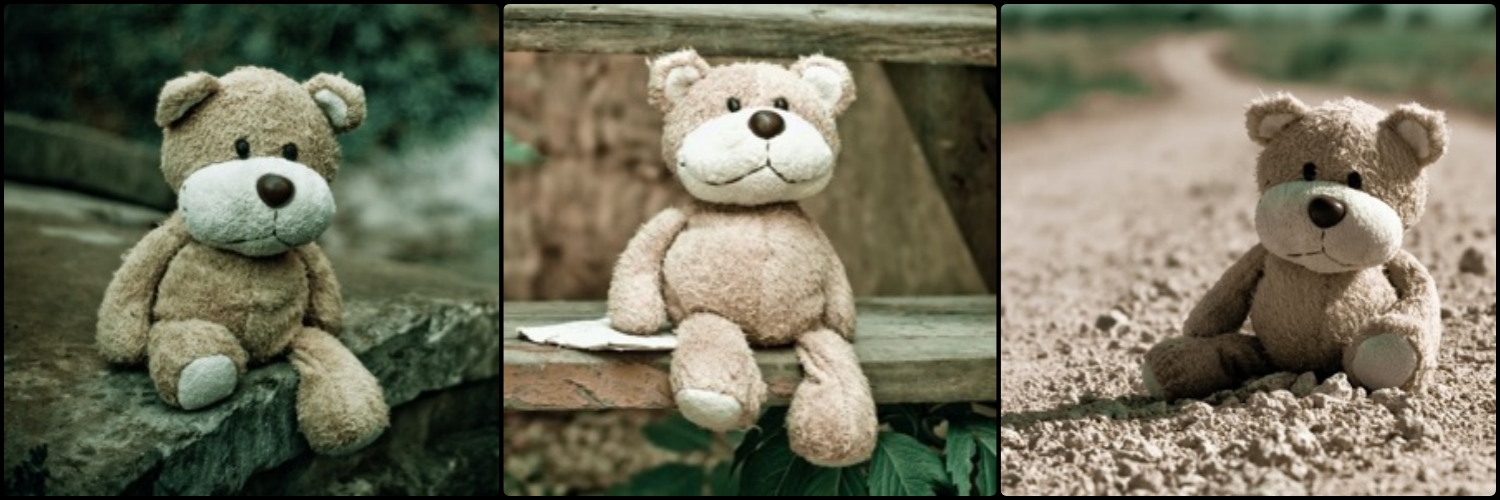}
\end{minipage}
&&
\begin{minipage}{0.47\columnwidth}
    \includegraphics[width=\linewidth]{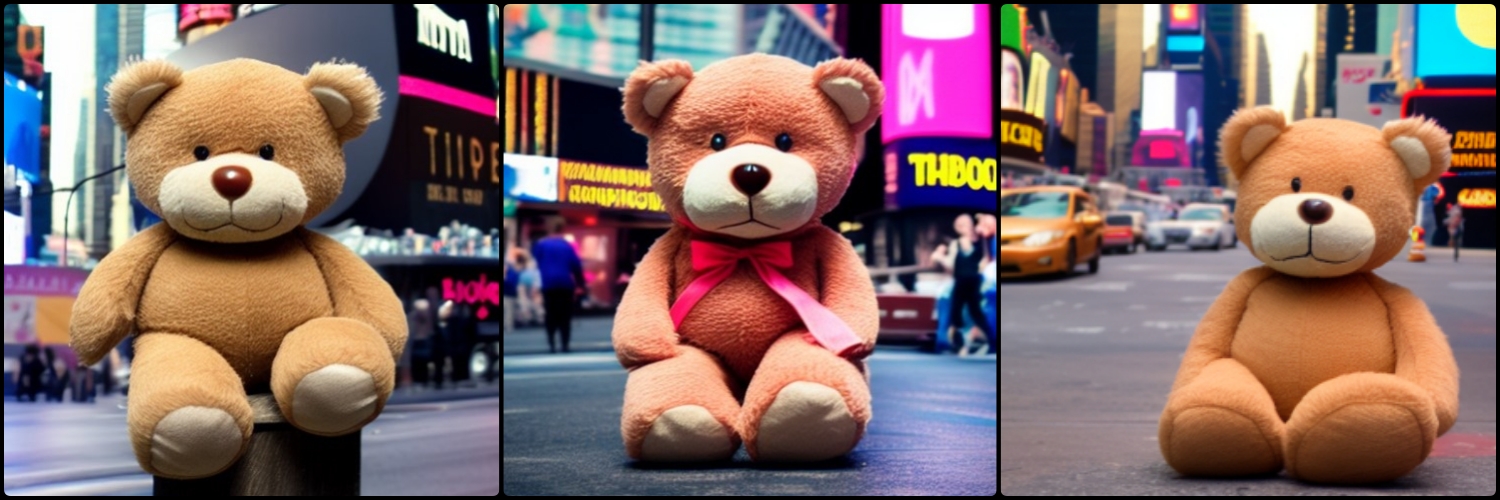}
\end{minipage}
&&
\begin{minipage}{0.47\columnwidth}
    \includegraphics[width=\linewidth]{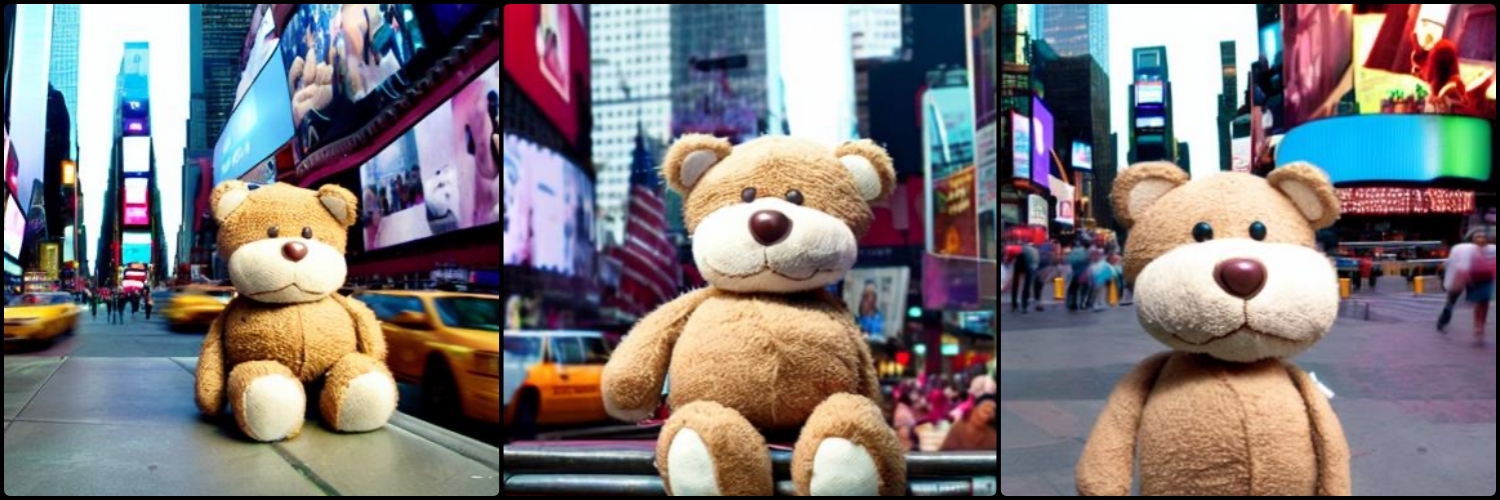}
\end{minipage}
&&
\begin{minipage}{0.47\columnwidth}
    \includegraphics[width=\linewidth]{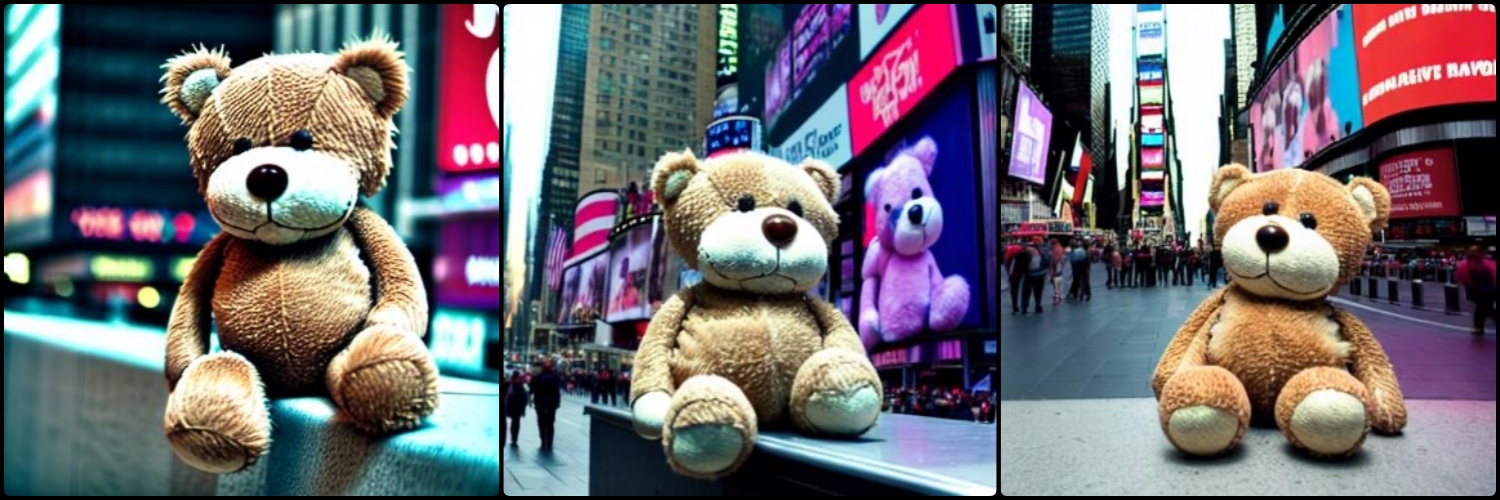}
\end{minipage}\\
&&&\multicolumn{9}{c}{\textbf{Scene change: V* teddy bear in Times Square}}\\
\toprule
&
\begin{minipage}{0.47\columnwidth}
    \includegraphics[width=\linewidth]{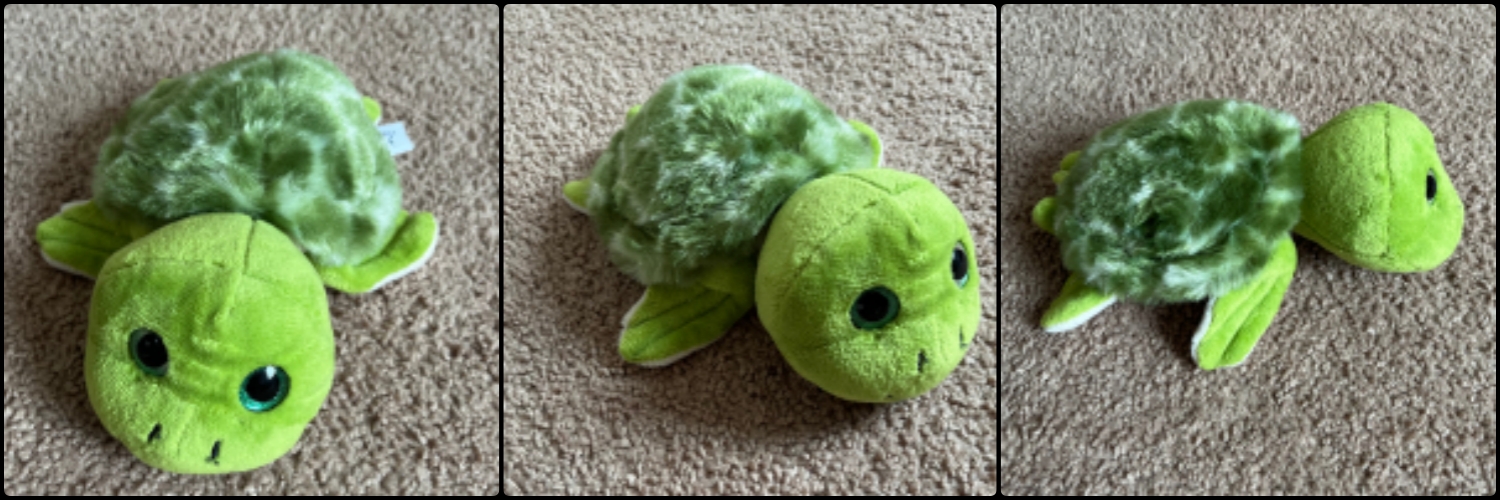}
\end{minipage}
&&
\begin{minipage}{0.47\columnwidth}
    \includegraphics[width=\linewidth]{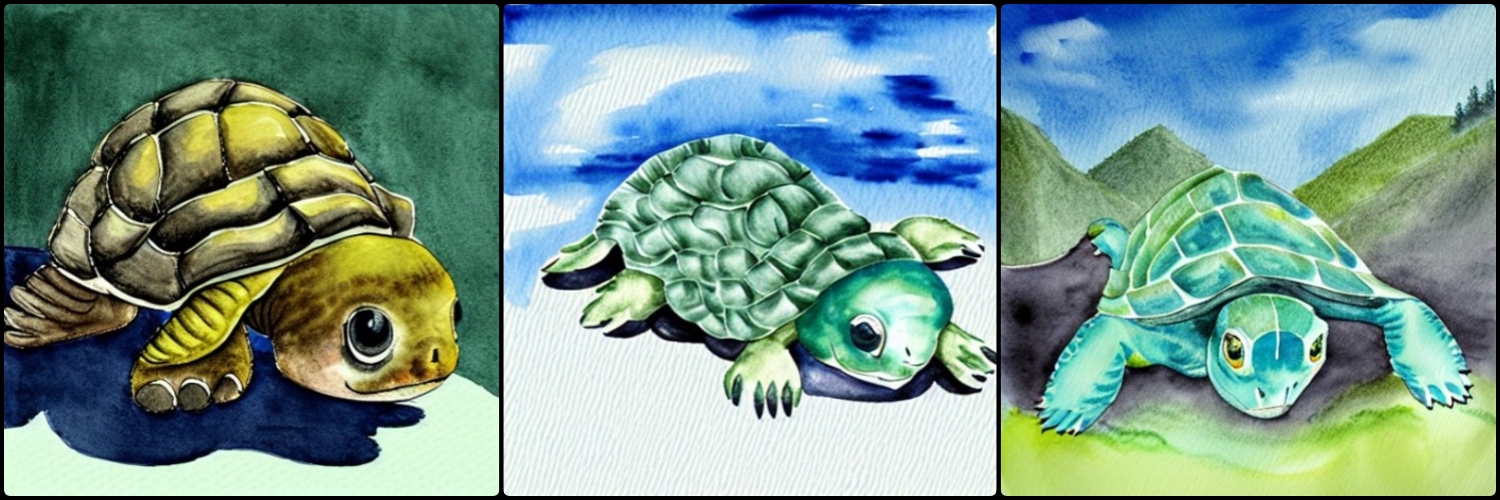}
\end{minipage}
&&
\begin{minipage}{0.47\columnwidth}
    \includegraphics[width=\linewidth]{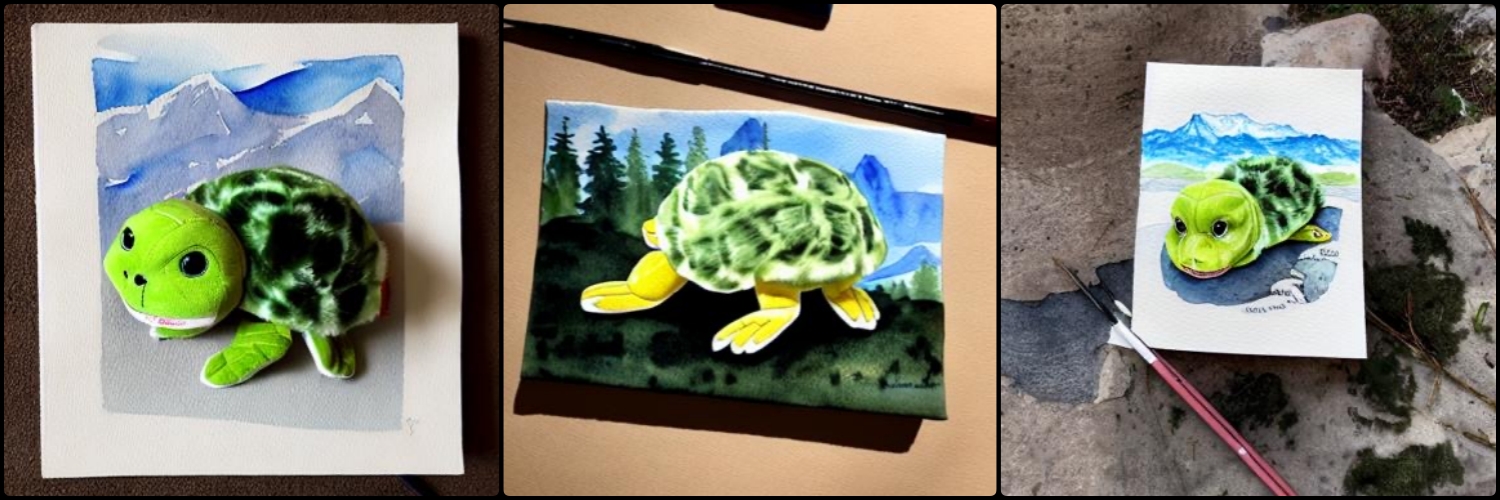}
\end{minipage}
&&
\begin{minipage}{0.47\columnwidth}
    \includegraphics[width=\linewidth]{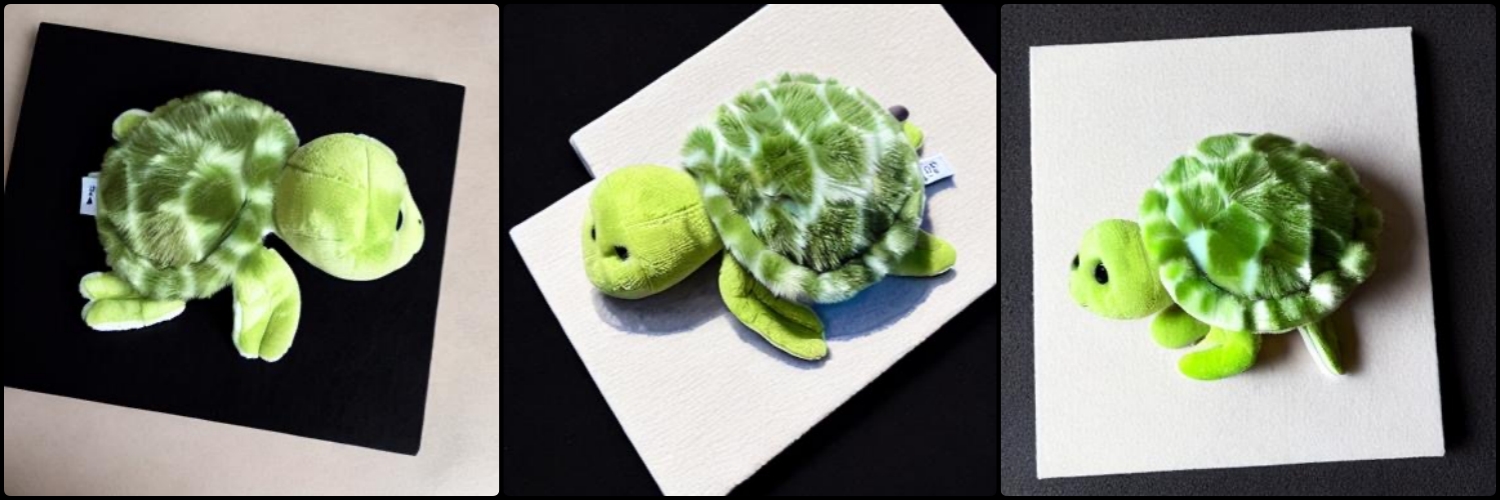}
\end{minipage}\\
&&&\multicolumn{9}{c}{\textbf{Artistic variations: A watercolor painting of V* tortoise plushy on a mountain}}\\
\toprule
&
\begin{minipage}{0.47\columnwidth}
    \includegraphics[width=\linewidth]{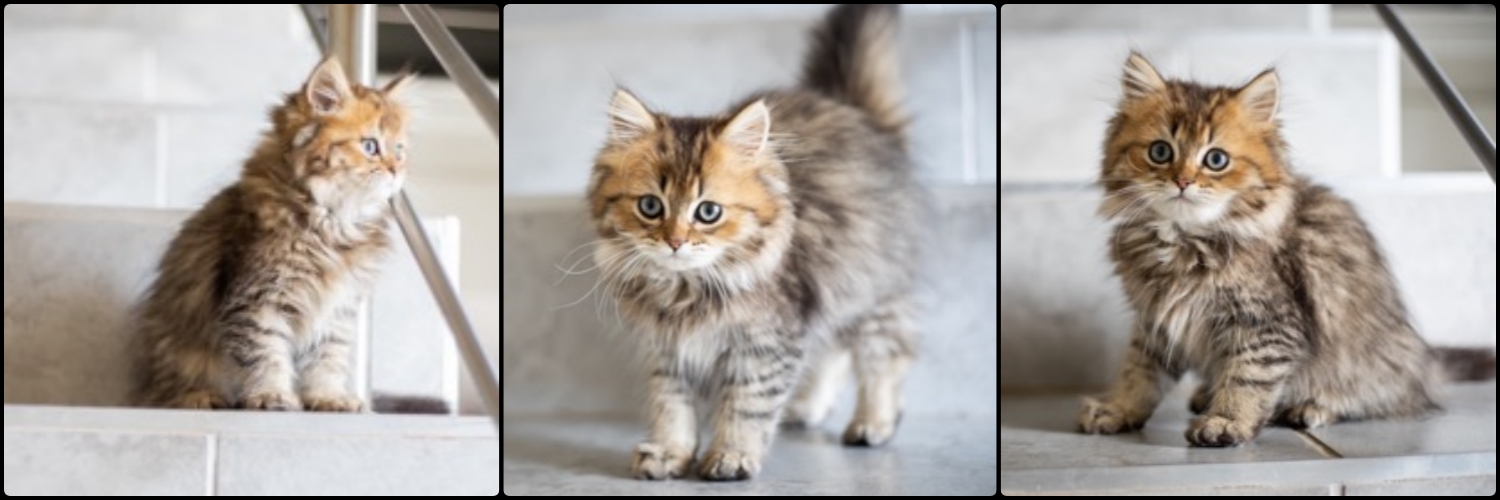}
\end{minipage}
&&
\begin{minipage}{0.47\columnwidth}
    \includegraphics[width=\linewidth]{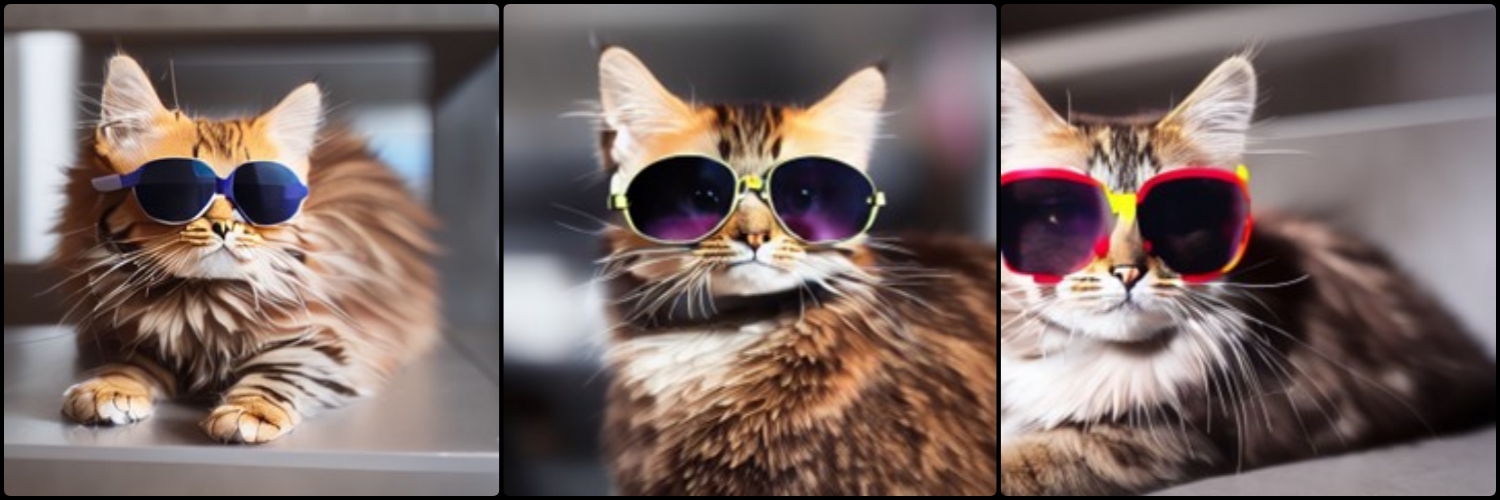}
\end{minipage}
&&
\begin{minipage}{0.47\columnwidth}
    \includegraphics[width=\linewidth]{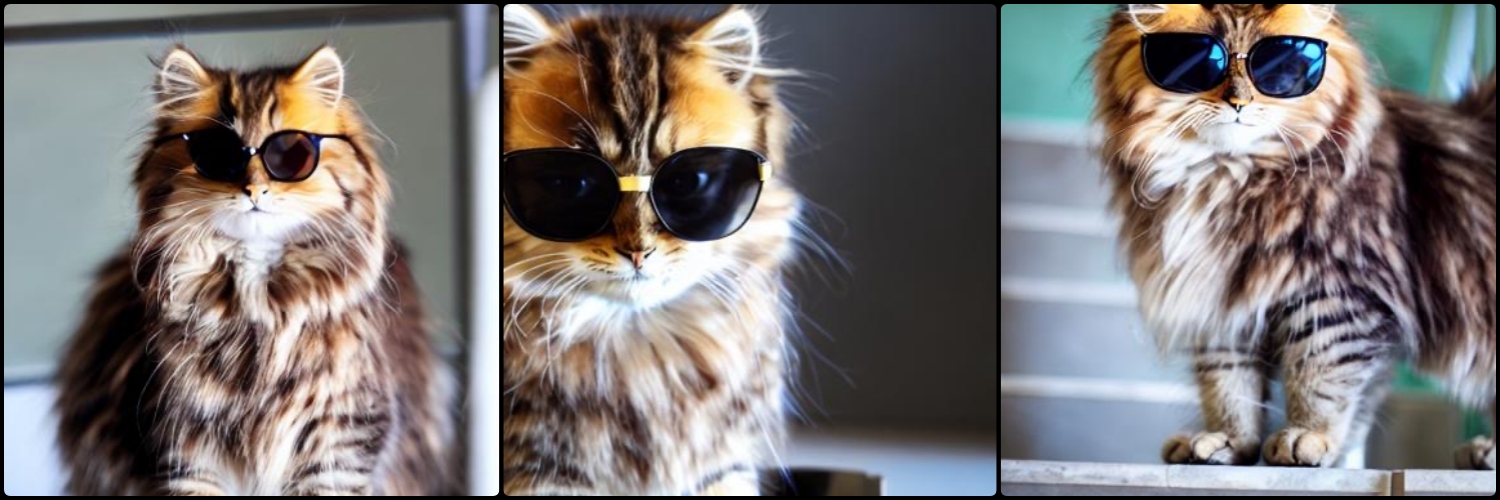}
\end{minipage}
&&
\begin{minipage}{0.47\columnwidth}
    \includegraphics[width=\linewidth]{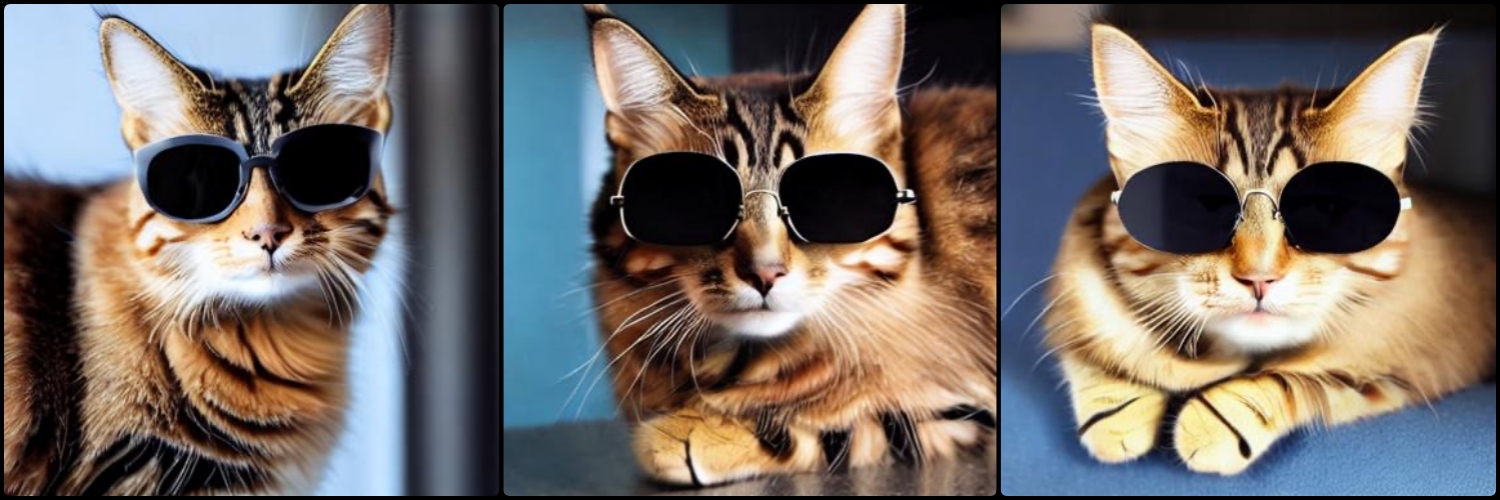}
\end{minipage}\\
&&&\multicolumn{9}{c}{\textbf{Accessorization: V* cat wearing sunglasses}}\\
\bottomrule
\end{tabular}
\caption{Fine-tuning results, where the results of CustomDiffusion$^\dag$ and DreamBooth$^\dag$ are from Custom Diffusion~\cite{kumari2022multi}.}
\label{fig:finetune-diff}
\end{figure*}


\textbf{Setting.} We fine-tune the \emph{cross-attention} layer of the pre-trained Stable Diffusion~\cite{rombach2022high} (model weights obtained from HuggingFace Hub\footnote{https://huggingface.co/CompVis/stable-diffusion-v1-4}~\cite{wolf2019huggingface}) by using our proposed low-rank MHA mechanism to enable the model to learn a new concept. 
For quantitative evaluation, we use the six objects included in the dataset released by CustomDiffusion~\cite{kumari2022multi} and one newly collected object, with the number of images contained in each one ranging from $4$ to $12$.

\textbf{Training Cost Comparison.} We first present the {training cost} required by all approaches in Table~\ref{tbl:diff-FID}. 
We train all the approaches on one Nvidia RTX A6000 GPU with the batch size as $1$ and the number of training steps as $500$.
Compared with CustomDiffusion and DreamBooth, our approach reduces the training time by $18.6\%$ and $61.6\%$, respectively (see Training Time in Table~\ref{tbl:diff-FID}), and decreases the extra storage space for each concept by $12.5\times$ and $1927.5\times$, respectively (see Extra Storage in Table~\ref{tbl:diff-FID}). The order of magnitude reduction of extra storage for each new concept is extremely important for the broad adoption of personalized text-to-image diffusion models, where users can prepare their diffusion models without the burden of model storage.

\textbf{Image Quality Comparison.} We then evaluate the quality of the synthesized images for all approaches.
We generate $20$ images for each learned target image (concept) by using the same text prompt for all approaches and calculate the FID~\cite{heusel2017gans,parmar2021buggy} between the synthesized and real images. The lower FID score indicates the smaller difference between the generated and the real images. As shown in Table~\ref{tbl:diff-FID}, our approach achieves the lowest FID than the existing works.
We further provide the qualitative comparison in Figure~\ref{fig:finetune-diff-gd}. In addition to generating the corresponding scenes accurately from text prompts, the \emph{V* Goldendoodle} generated by our method is the closest to the real image, while the \emph{V* Goldendoodle} synthesized by DreamBooth~\cite{ruiz2022dreambooth} has obvious differences from the real one, and images synthesized from CustomDiffusion~\cite{kumari2022multi} contain less natural mouth as it has been deformed. In Figure \ref{fig:finetune-diff}, we further show that even with much fewer computation resources, \emph{i.e.,}  less training time and fewer number of GPUs for model fine-tuning, we can still generate high-quality images.

\textbf{MS-COCO Evaluation.} Lastly, we perform the experiments to understand if the fine-tuned models can generate images that are unrelated to the learned target subject (\emph{V*}). We use the prompted text of $5,000$ images from the MS-COCO 2017~\cite{lin2014microsoft} validation set to generate images and calculate the FID. As shown in Table~\ref{tbl:diff-FID-mscoco}, the FID from the personalized models are similar to the pre-trained text-to-image model, indicating that they can synthesize high-quality images for unrelated concepts. Thus, the model fine-tuned by our method still holds the distribution of the synthesized images as the pre-trained model.
\section{Conclusion}

This paper fundamentally investigates the low-rankness in the multi-ahead attention layer of the emerging vision models and proposes that the head-level low-rankness should be explored for efficient model design, bringing highly efficient low-rank ViT compression solution. Our method not only outperforms existing compression approaches by providing higher performance but also brings faster practical speedup. Particularly, our finding is further applied for efficient customization of text-to-image diffusion models, outperforming the state-of-the-art solutions.

\section{Acknowledgements}
This work was partially supported by National Science Foundation under Grant CCF-1937403 and CCF-1955909.

\clearpage
\nocite{langley00}

\bibliography{icml2023}
\bibliographystyle{icml2023}

\newpage
\appendix
\onecolumn

\end{document}